\documentclass[10pt,journal]{IEEEtran}

\usepackage{graphicx}

\IEEEoverridecommandlockouts

\graphicspath{{./figures/}}

\usepackage{cite}

\usepackage{wrapfig}

\usepackage{algorithm}
\usepackage{amssymb}
\usepackage{amsmath}
\usepackage{graphicx}
\usepackage{url}

\newtheorem{proposition}{Proposition}
\newtheorem{theorem}{Theorem}
\newtheorem{remark}{Remark}
\newtheorem{lemma}{Lemma}

\newtheorem{corollary}{Corollary}

\usepackage{amsmath}
\usepackage{graphicx}
\usepackage{url}
\usepackage{xcolor}
\usepackage{epstopdf}

\title{\huge Scale-free vision-based aerial control of a ground formation with hybrid topology}

\author{Miguel Aranda, Youcef Mezouar, Gonzalo L\'{o}pez-Nicol\'{a}s and Carlos Sag\"{u}\'{e}s
\thanks{M. Aranda and Y. Mezouar are with Universit\'{e} Clermont Auvergne, CNRS, SIGMA Clermont, Institut Pascal, F-63000 Clermont-Ferrand, France. {\tt\footnotesize first\_name.last\_name@sigma-clermont.fr}}
\thanks{G. L\'{o}pez-Nicol\'{a}s and C. Sag\"{u}\'{e}s are with Instituto de Investigaci\'{o}n en Ingenier\'{i}a de Arag\'{o}n, Universidad de Zaragoza, Zaragoza, Spain. {\tt\footnotesize gonlopez@unizar.es, csagues@unizar.es}}
\thanks{This work was supported by the French Government via programs FUI (project Aerostrip) and Investissements d'Avenir (I-SITE project CAP 20-25 - MaRoC), and by the Spanish Government/European Union through project DPI2015-69376-R.}
  \thanks{\textcolor{red}{This is the accepted version of the manuscript: M. Aranda, Y. Mezouar, G. López-Nicolás and C. Sagüés, "Scale-Free Vision-Based Aerial Control of a Ground Formation With Hybrid Topology," in IEEE Transactions on Control Systems Technology, vol. 27, no. 4, pp. 1703-1711, July 2019, doi: 10.1109/TCST.2018.2834308. 
\textbf{Please cite the publisher's version}. For the publisher's version and full citation details see:\\
\protect\url{https://doi.org/10.1109/TCST.2018.2834308}. 
}}

 \thanks{© 2019 IEEE.  Personal use of this material is permitted.  Permission from IEEE must be obtained for all other uses, in any current or future media, including reprinting/republishing this material for advertising or promotional purposes, creating new collective works, for resale or redistribution to servers or lists, or reuse of any copyrighted component of this work in other works.}
 }
\begin{document}
\maketitle
\thispagestyle{headings}
\pagestyle{headings}
\begin{abstract}
We present a novel vision-based control method to make a group of ground mobile robots achieve a specified formation shape with unspecified size. Our approach uses multiple aerial control units equipped with downward-facing cameras, each observing a partial subset of the multirobot team. The units compute the control commands from the ground robots' image projections, using neither calibration nor scene scale information, and transmit them to the robots. The control strategy relies on the calculation of image similarity transformations, and we show it to be asymptotically stable if the overlaps between the subsets of controlled robots satisfy certain conditions. The presence of the supervisory units, which coordinate their motions to guarantee a correct control performance, gives rise to a hybrid system topology. All in all, the proposed system provides relevant practical advantages in simplicity and flexibility. Within the problem of controlling a team shape, our contribution lies in addressing several simultaneous challenges: the controller needs only partial information of the robotic group, does not use distance measurements or global reference frames, is designed for unicycle agents, and can accommodate topology changes. We present illustrative simulation results.
\end{abstract}
\section{Introduction}
Compared to single-robot setups, multirobot systems provide increased efficiency and reliability, which makes them a very popular research subject. We address here the particular problem of formation shape control, where multiple robots move to collectively form a desired shape with unspecified location, rotation, and size \cite{Mesbahi10}. As opposed to team formations defined as rigid, fixed-size patterns of agent positions \cite{Anderson08,Cortes09,Ren10,Cai15,Oh15}, we study here formations specified only by angular constraints, also known in the literature as \textit{bearing formations}. Controlling them is a problem of current interest \cite{Bishop11,Eren12,Franchi12,Coogan12,Schoof14,Lin14,Zhao15,ZelazoCDC15,Han15}, crucial in any application scenario where only angular measurements are available. Also, such shape control allows, e.g., to regroup the agents in an organized manner before addressing a subsequent task, create desired vicinities (with no specific regard for distances) between concrete agents, or form a certain favorable shape as fast as possible, e.g., to react to a threat. Here, we investigate this relevant problem considering an infrastructure-free scenario, in the sense that the proposed robotic system does not depend on any elements external to itself (e.g., GPS or motion capture data), which is interesting in practice for flexibility and robustness. To this end, the use of vision sensors is quite appealing. Cameras naturally lend themselves to angle-based control, and enable various multirobot behaviors, e.g., coordinated motion \cite{Moshtagh09} and orientation alignment \cite{Montijano13}. Aerial vision has been identified as particularly interesting for control and environment perception tasks in robotics \cite{NMichael08,Lacroix2011,SchwagerProcIEEE11Cameras,Lopez-Nicolas12,GarciaCarrillo14,ArandaTRO15,Poiesi2015,Chen2016}, due to
cameras being low-weight sensors that provide very rich data.

\vspace{0pt}

\begin{figure}
\centerline{
\includegraphics[trim = 0.7cm 0.8cm 0.7cm
0cm,width=0.935\linewidth]{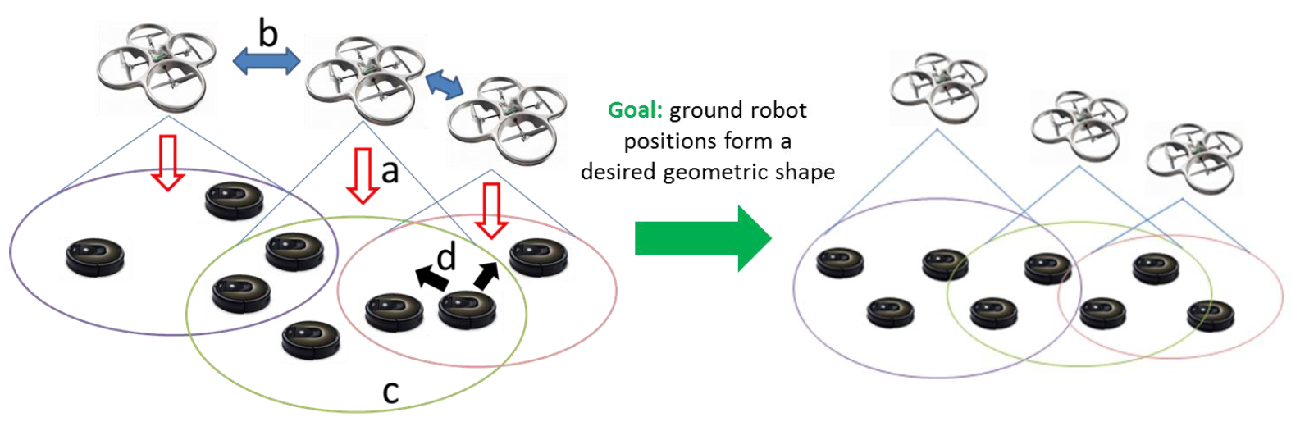}}
\caption{Overview of the multirobot control system. Multiple (three,
in this example) moving aerial units are used. Each computes
(Section \ref{ssimila}) and transmits (a) motion commands for a set
of ground robots in its camera's field of view (c). The robots that
are controlled by multiple cameras combine the multiple received
commands (d) to obtain their motion input, as described in Section
\ref{smulti}. The UAVs can communicate (b) to coordinate their
actions. The control task is for the ground robots' positions to
form a specified shape.}\label{fig_esquemamulticam}
\end{figure}

\noindent \textbf{Properties of the aerial vision-based framework:} We build here on a control framework of
ground formations with specified size that we presented in \cite{Lopez-Nicolas12,ArandaTRO15}. Our method is
based on a two-layer architecture where a set of downward-facing cameras onboard Unmanned Aerial
Vehicles (UAVs) are used to observe and control the ground robots. The system setup proposed is
illustrated in Fig. \ref{fig_esquemamulticam}. The aerial units detect and identify the robots, and measure their position and heading, using image information. They compute a similarity from their current image and a template image (which encodes the desired shape) to define the motion goals for the ground robots. Crucially, each UAV controls only a partial \textit{subset} of the robots, and uses solely uncalibrated image information. No common reference frame among UAVs is needed, and they can displace and rotate while hovering throughout execution without affecting the control convergence. These prominent practical advantages facilitate simple, robust and flexible implementation (see Section \ref{sdiscussion}). We require certain overlaps between subsets, and establish how a ground robot receiving multiple commands integrates them to compute its movement. Via Lyapunov-based analysis, we show that the proposed controller makes the team asymptotically reach the prescribed shape. We also provide a method for the UAVs to control their motions to appropriately cover the ground agents. In terms of \textit{topology} (i.e., the interactions between elements of the multiagent system), our framework is \textit{hybrid}; this means that it is neither centralized (there are multiple UAVs, each handling partial information) nor purely distributed, because each UAV acts as a central node for a subset of robots.

\vspace{0pt}

\noindent \textbf{Related literature on formation shape control:} The literature on non-centralized control of bearing formations requires each agent to satisfy desired angular constraints with respect to a subset of the other agents. To guarantee the achievement of the prescribed team shape, the interaction graph that encapsulates the system's topology must satisfy \textit{parallel (or bearing) rigidity} conditions. Within this framework, \cite{Eren12,Franchi12,Schoof14} present distributed control laws for these formations relying only on angular measurements, while in \cite{Bishop11} distances are also used. The work \cite{Zhao15} uses only bearings and requires the robots to synchronize their orientations during execution. All these approaches need the relative angles used by the agents to be expressed in a common orientation reference, contrary to our method, where the measurements are expressed in the different and independent image frames of the multiple cameras. Importantly, in these related methods the final team shape has a constrained orientation in the workspace. For numerous applications (e.g., team navigation in formation), a pattern with no constraints on its orientation --as allowed by our method-- is more flexible and efficient. The controller in \cite{ZelazoCDC15} stems from principles of $SE(2)$ rigidity theory. Each agent uses locally expressed bearings and the relative orientation of neighboring agents' frames. The scheme in \cite{Lin14} employs a topological representation via the complex graph Laplacian, and also controls the team shape without global references, albeit using both angles and distances. Similar information is assumed in \cite{Coogan12} and \cite{Han15}, which deal with formations of adjustable size. The topology of the system may change over time, and the study of such changes is a prevalent topic \cite{Olfati07,Liu15jfi}. Unlike in the works cited above, here we investigate this \textit{switching topology} scenario, and we assume the ground robots have nonholonomic (unicycle) kinematics.

\vspace{0pt}

\noindent \textbf{Statement of contributions:} Relative to the literature on formation shape control, our contribution lies in that we consider more challenging conditions. Specifically, our non-centralized, partial information-based method requires no global reference frames and relies solely on pixel image information and no range measurements. We consider unicycle agents, and study the stability under switching topologies. Also, our proposed hybrid architecture represents a novel perspective on the problem, with practical advantages in, e.g., ground robot simplicity, task supervision and
flexibility, discussed in Section \ref{sdiscussion}.

Our previous method proposed in \cite{ArandaTRO15} also considers a two-layer framework and a similar scenario, controlling a fixed-size ground formation with cameras that can perform team scale adjustments using supplementary information. In this paper we present novel contributions relative to that work: \\
$\bullet$ Here, the size of the obtained formation is flexible, a property that cannot be achieved with the method in \cite{ArandaTRO15}, and that is significant and interesting in its own right. It can, e.g., increase the motion efficiency and reduce the task execution time. \\
$\bullet$ The approach has significantly lower information requirements. The aerial units use only pixel coordinates of the images of the ground robots, and no information of absolute scale. Camera calibration, metric data or image scale estimates obtained with supplementary information are not needed. In contrast, \cite{ArandaTRO15} requires some of these sources of information to obtain continuous knowledge of the scale of the imaged scene. \\
$\bullet$ In contrast with \cite{ArandaTRO15}, here we provide formal stability guarantees under switching topologies, which is an important aspect as topology switches will typically occur in practice.
\section{System description and operating assumptions}
\label{ssysdes}
Let us describe the characteristics of the proposed system and the conditions of operation that are assumed. \\
\textbf{Task:} The positions of a set, $S$, of $n$ mobile robots must attain a prescribed shape, with unspecified size. \\
\noindent \textbf{Architecture:} The system has a two-layer architecture: the ground robot layer, and a set of $m$ UAVs that
observe the robots and control their motion.\\
\textbf{Vehicle dynamics:} Each UAV remains near-hovering (i.e., its yaw axis is maintained vertical) and its translation is commanded via kinematic control (as, e.g., in \cite{Bourquardez09}; see details in Section \ref{scoordaerial}). The ground robots have unicycle kinematics and move on a horizontal ground plane.\\
\textbf{Perception:} Each UAV carries a fixed perspective camera facing downwards. Using vision processing, it can detect and identify those ground robots in its field of view, and compute their image positions and headings. The ground robots do not require any sensors for the task addressed. \\
\textbf{Prior information:} Each aerial unit knows the prescribed formation shape with the identification of each robot, represented in the form of a \textit{template image} (Section \ref{ssimila}). \\
\textbf{Communications:} Each UAV sends commands via wireless communication to robots in its camera's field of view. Two UAVs that observe robots in common at any time communicate via wireless to coordinate their motions/actions (see ``Coordination and control" below), and multi-hop exchanges may also be used. The ground robots do not transmit any data. \\
\noindent \textbf{Topology:} The control topology is hybrid (not centralized, not purely distributed). For system stability purposes, we characterize it as follows. Each UAV views a subset of robots $S_{j} \subseteq S$, $j=1,...,m$, and controls a set $S_{j}^c \subseteq S_{j}$, being these sets time-varying. Formation achievement requires overlaps between these sets. We define $\mathcal{G}_c$ as an undirected graph where each node is a UAV and there is an edge $(j,k)$ if $card(S_{j}^c \cap S_{k}^c) \geq 2$ (i.e., the two UAVs are controlling at least two common robots). We define \textit{topological conditions}:
\begin{itemize}
\item  TC1: $\mathcal{G}_c$ is connected.
\item TC2: $\bigcup_{j=1,...,m} S^{c}_j=S$ (i.e., every robot is controlled).
\item  TC3: $card(S_{j}^c \bigcap S_{k}^c)=2$ if $j$, $k$ are neighbors in $\mathcal{G}_{c}$, $0$ otherwise (i.e., neighboring UAVs share exactly two robots, and non neighboring UAVs share no robots).
\item  TC4: $S_{j}^c \bigcap S_{k}^c  \bigcap S_{l}^c =\emptyset$ if $j$, $k$, $l$ are all different. That is, the intersections between the sets are mutually disjoint.
\item  TC5: $\mathcal{G}_c$ is a tree.
\end{itemize}
We define $\mathcal{P}$ as the set of all possible topologies that satisfy TC1, TC2, TC3, TC4 and TC5, and $\mathcal{Q}$ as the set of those topologies that satisfy TC1 and TC2. We denote the system's topology at time $t$ by $p(t)$. We assume $p(t=0) \in \{\mathcal{P} \cup \mathcal{Q}\}$. \\
\textbf{Coordination and control.} \textit{Ground robots:} They integrate and follow the motion commands received from the UAVs (Section \ref{smulti}). They do not sense/communicate with one another, but move in coordination thanks to the aerial units. \textit{Aerial units:} Each unit $j$ sends a motion objective, and image distance information --obtained as shown in Section \ref{smulti}-- to each robot in $S_{j}^c$. Each UAV coordinates its motion and the definition of $S_{j}$ and $S_{j}^c$ with its neighbors (in terms of the graph $\mathcal{G}_c$) to maintain $p(t)$ within the set $\mathcal{P}$ (Section \ref{scoordaerial}). They also ensure each topology is active for a lower bounded time span.
\begin{figure}
\centerline{
\includegraphics[trim=5cm 6.6cm 5cm 3cm,width=0.57\linewidth]{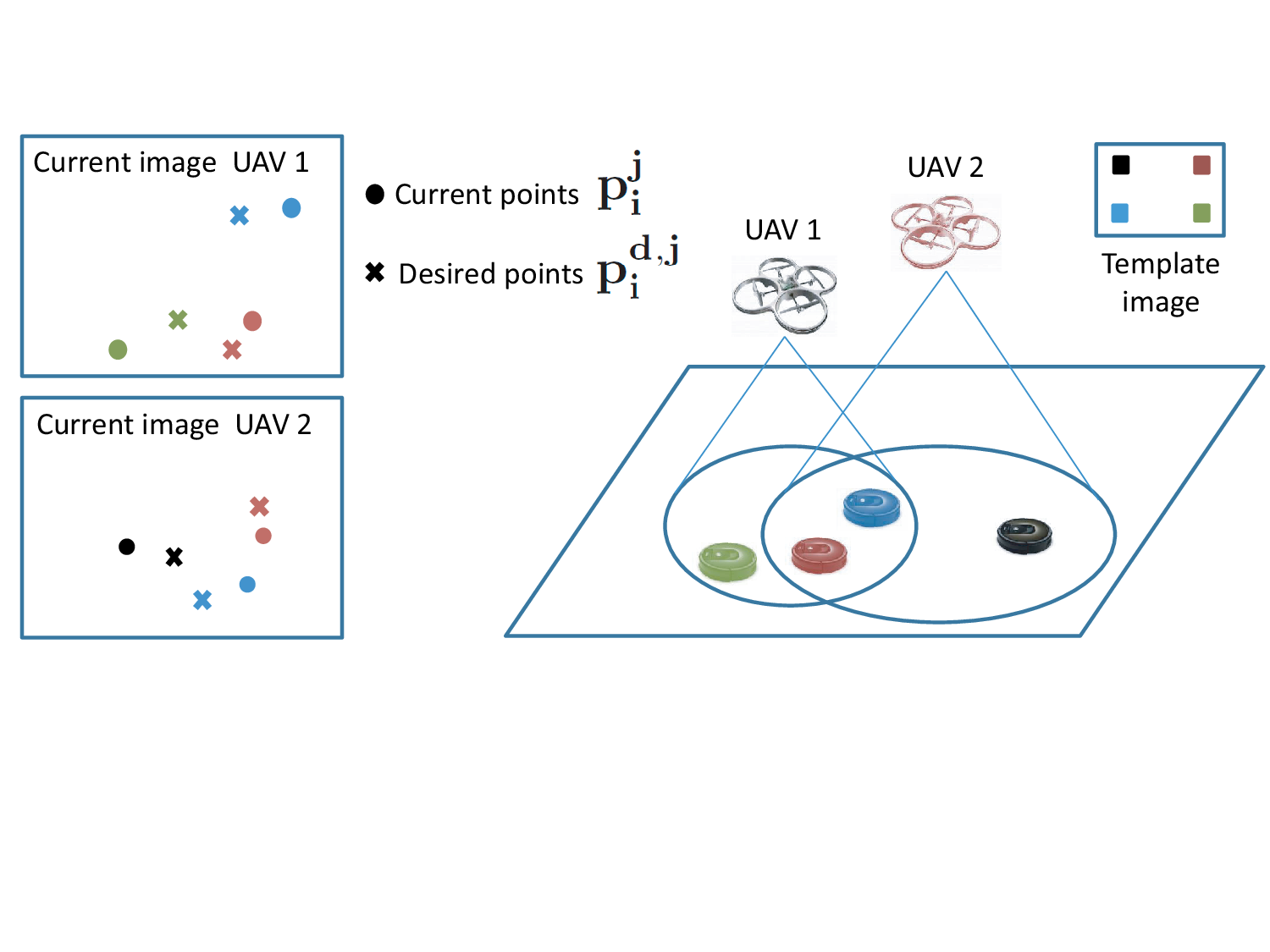}
}\caption{Each UAV (i.e., 1 or 2) sees and controls only a partial subset of robots, using the corresponding partial set of template and current image points to compute the desired image points via a least-squares similarity.}\label{figuavssimil}
\end{figure}
\section{Similarity-based motion goals computation}
\label{ssimila}
Consider next a given control unit $j$. To compute the control goals for the ground robots, it uses two perspective images: \\
$\bullet$ The \textit{template image}, which is a predefined, fixed top view, with arbitrary scale, of the desired formation shape. Each robot is represented by a point $\mathbf{p_{i}'}$, in pixel coordinates. Unit $j$ only uses those points in the template image that correspond to robots $j$ controls (i.e., $i \in S_{j}^{c}$).\\
$\bullet$ The \textit{current image}, which is a top view of the current configuration of the subset of robots $i \in S_{j}^c$, each of which is represented by a point, $\mathbf{p_{i}^{j}}$, in pixels. \\
We propose a strategy where each camera uses the template and current image points to compute a \textit{similarity} transformation (translation, rotation, and scaling) that aligns them with least-squares error \cite{Gower04}. The 2D similarity we calculate relating the two point sets is parameterized as follows:
\begin{align}
{\bf H_{s}^{j}}
 = \left[ \begin{array}{cc} s_{j}\cos\phi_{j} & -s_{j}\sin\phi_{j} \\
 s_{j}\sin\phi_{j} & s_{j}\cos\phi_{j}\end{array}  \right],
\label{eHeuclideanl}
\end{align}
and encodes the rotation of the template shape by $\phi_{j} \in [-\pi, \pi)$ and its scaling by $s_{j} \in \mathbb{R}^{+}$. This, together with a translation such that the current points' centroid is maintained, is used to obtain what we call the \textit{desired} points, $\mathbf{p^{d,j}_{i}}$, which define the robots' motion goals. Algorithm \ref{algorithmcam} summarizes the process and Fig. \ref{figuavssimil} illustrates it for two different UAVs. A key decoupling between camera motion and ground control is expressed next.
\begin{algorithm}
  \begin{enumerate}
    \item Select the points $\mathbf{p_{i}'}$ for $i \in S_{j}^c$ from the \textit{template image} and (if required) translate them to make their centroid zero, obtaining the set of points $\mathbf{p'^{j}_{ic}}$.
    \item \textbf{While} control executes \textbf{do}:
      \begin{enumerate}
    \item Acquire a new \textit{current image}.
    \item Detect and identify in the current image the points $\mathbf{p_{i}^{j}}$ corresponding with the current robot positions.
    \item Subtract the centroid, $\mathbf{c_{p}^{j}}$, of the points $\mathbf{p_{i}^{j}}$, to create a new set of points $\mathbf{p_{ic}^{j}}$ with zero centroid.
    \item Compute the similarity $\mathbf{H_{s}^{j}}$ that, applied on $\mathbf{p'^{j}_{ic}}$, aligns them with $\mathbf{p_{ic}^{j}}$ with least-squares error \cite{Gower04}.
    \item Compute the desired image points, expressed in the current image, as: ${\bf p^{d,j}_{i}}={\bf{H_{s}^{j}}}$ $\mathbf{p'^{j}_{ic}}$ $+\bf{c_{p}^{j}}$.
    \end{enumerate}
    \end{enumerate}
\caption{Computation by camera $j$ at each time instant of the desired image positions for the robots $i \in S_{j}^c$} \label{algorithmcam}
\end{algorithm}
\noindent \textbf{Property 1:} The ground positions associated with the desired points, defined from the optimal similarity by a given camera, are invariant to the downward-facing camera's position, orientation and calibration.
\begin{IEEEproof} Consider two arbitrary configurations for camera $j$: $j_a$ and $j_b$. These can be linked by a similarity $\mathbf{G_{ba}}$, so $\mathbf{p_{ic}^{j_b}}=\mathbf{G_{ba}}\mathbf{p_{ic}^{j_{a}}}$ for $i \in S_{j}^c$. Obviously, $\mathbf{p'^{j_{b}}_{ic}}=\mathbf{p'^{j_{a}}_{ic}}$. The similarity (\ref{eHeuclideanl}) can be obtained for $k=a$ or $b$ solving via least-squares a linear system with equations: $\mathbf{H_{s}^{j_k}} \mathbf{p'^{j_{k}}_{ic}}-\mathbf{p^{j_{k}}_{ic}}=\mathbf{0}$ $\forall i \in S_{j}^c$. Comparing the two systems, we have $\mathbf{H_{s}^{j_b}}=\mathbf{G_{ba}}\mathbf{H_{s}^{j_a}}$. Then (see Algorithm \ref{algorithmcam}) $\mathbf{p^{d,j_{b}}_{i}}-\mathbf{c_{p}^{j_{b}}}=\mathbf{H_{s}^{j_{b}}}\mathbf{p'^{j_{b}}_{ic}} =\mathbf{G_{ba}}\mathbf{H_{s}^{j_a}}\mathbf{p'^{j_{a}}_{ic}}=\mathbf{G_{ba}}({\bf p^{d,j_{a}}_{i}}-\mathbf{c_{p}^{j_{a}}})$, where clearly $\mathbf{c_{p}^{j_{a}}}$ and $\mathbf{c_{p}^{j_{b}}}$ are the centroids of the desired point sets. Hence, the desired points $i$ for $a$ and $b$ are projections of the same ground position $\forall i \in S_{j}^c$.
\end{IEEEproof}
If $\bf {p_{i}^{j}}=\bf {p^{d,j}_{i}}$ $\forall i \in S_{j}^c$, clearly, the robots in this \textit{subset} form the desired \textit{sub-shape}. Our control goal is thus to move them so that they meet this condition. Note, however, the important challenges we face: these sub-shapes must fit together in the \textit{full} formation (\textit{overlaps} between subsets are needed), and a robot can receive \textit{multiple partial and inconsistent} motion goals --see, e.g., the two robots in the intersection in Fig. \ref{figuavssimil}--. The following section describes how these issues are solved.
\section{Coordinated ground robot control scheme}
\label{smulti}
We explain next how a control unit $j$ computes the information to be sent to a controlled robot $i$. Given a camera with usual characteristics, we can define a scale $r_{j}>0$ (in \textit{pixels/m} units) relating $j's$ image distances with the metric distances between ground entities. This scale is unknown, freely time-varying, and different for each camera.

\begin{figure}
\centerline{
\includegraphics[trim = 0cm 2.5cm 0cm 2.5cm,width=0.82\linewidth]{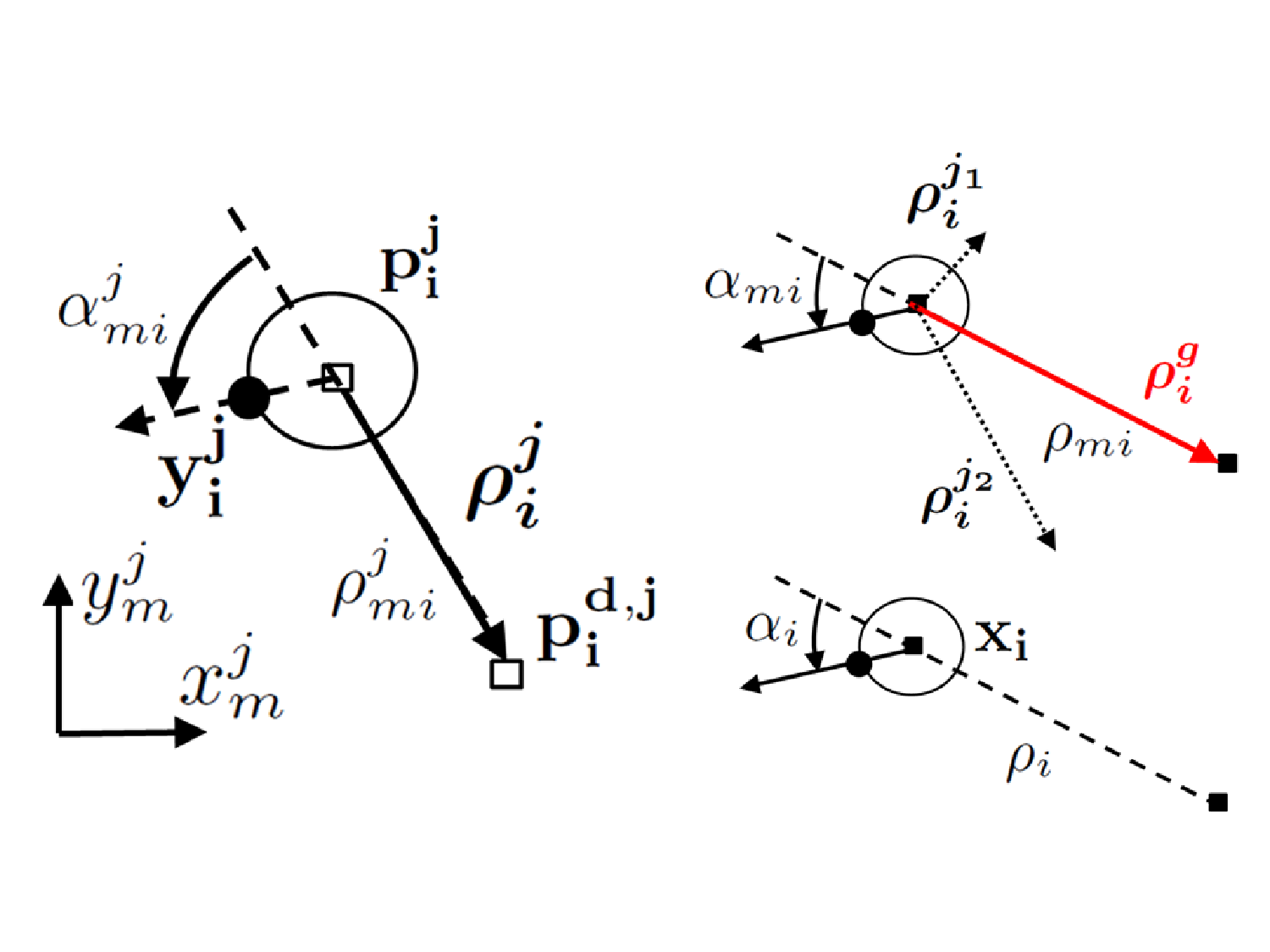}
}\caption{Left: geometric variables and control vector computed for robot $i$ by camera $j$, defined in its image. Right-top: representation of $i's$ global motion vector computed from image information received from two cameras $j_1$ and $j_2$. Right-bottom: state of the robot on the ground plane.}\label{figMHimageplane}
\end{figure}

The parameters of the control scheme are depicted in Fig. \ref{figMHimageplane}. Using the strategy described in the previous section, we can define $\mathbf{p^{d,j}_i}$ in the current image and compute the vector:
\begin{equation}
\boldsymbol{\rho_{i}^{j}}=\mathbf{p_{i}^{d,j}} - \mathbf{p_{i}^j}.
\label{erhoij}
\end{equation}
By detecting the robots' positions and orientations in the images captured by its onboard camera, the control unit can obtain $\rho_{mi}^j = || \mathbf{p^{d,j}_{i}} - \mathbf{p^{j}_{i}} ||$, and it can also define the unit vector $\widehat{\boldsymbol{\rho}}_{\boldsymbol{i}}^{\boldsymbol{j}} =\boldsymbol{\rho_{i}^{j}}/\rho_{mi}^{j}$ and the unit vector in the direction of the robot's heading (which $j$ measures in the image) $\mathbf{y^{j}_{i}}$ . Then, the control unit can compute the following angular parameter:
\begin{eqnarray}
\alpha_{mi}^{j} = atan2 \left( - \left[     \left( \begin{array}{cc} \widehat{\boldsymbol{\rho}}_{\boldsymbol{i}}^{\boldsymbol{j}}  \\ 0 \end{array} \right)\times \left(\begin{array}{cc} \mathbf{y^{j}_{i}} \\ 0 \end{array} \right)\right]_{z}, -\widehat{\boldsymbol{\rho}}_{\boldsymbol{i}}^{\boldsymbol{j}^T}\mathbf{y^{j}_{i}} \right),
\end{eqnarray}
\noindent where $[\cdot]_{z}$ denotes the $z$-axis coordinate. From $\rho_{mi}^{j}$ and $\alpha_{mi}^{j}$, unit $j$ obtains $\boldsymbol{\rho_{i}^{j}}$  expressed in $i's$ frame, and sends it to $i$.
\noindent \textbf{Combination of multi-camera commands:} Robot $i$ may receive from multiple cameras simultaneous control goals that are inconsistent: each vector $\boldsymbol{\rho_{i}^{j}}$ for different $j$ is computed from a different subset of robots --so it will point in a different direction--, and is also associated with a different scale $r_{j}$.

To solve the scale inconsistency, unit $j$ sends robot $i$ the identification of the robots that are closest to it in $j's$ image (i.e., its physical nearest neighbors, in any direction around the robot), and the value of the image distances between $i$ and each of these robots (i.e., $||\mathbf{p^{j}_{i}}-\mathbf{p^{j}_{i_{o}}}||$ for a given neighbor $i_{o}$). From TC3, robot $i$ can compute the scale ratio between all cameras it receives data from: assume aerial unit $k$ also sends data to robot $i$, and that a robot $i_{o}$ is a physical neighbor of $i$, Both $j$ and $k$ view $i$ and $i_{o}$. Then, $i$ can compute the relative scale as follows: $r_{kj}=r_{k}/r_{j}=||\mathbf{p^{k}_{i}}-\mathbf{p^{k}_{i_{o}}}|| / ||\mathbf{p^{j}_{i}}-\mathbf{p^{j}_{i_{o}}}||$.

We define the global motion vector that robot $i$ computes as a weighted sum that integrates all its motion goals:
\begin{equation}
\boldsymbol{\rho_{i}^{g}}=\frac{1}{card(C_{i})} \sum_{j \in C_{i}} \left[ \sum_{k \in C_{i}} r_{kj} \boldsymbol{\rho_{i}^{j}} \right].
\label{eqrig}
\end{equation}
\noindent where $C_i$ is the set of indexes of the UAVs that send commands to $i$. We show next how this achieves scale consistency. Denote as $\bf {x^{d,j}_{i}}$ and $\bf {x_{i}}$ the ground positions associated
with $\bf {p^{d,j}_{i}}$ and $\bf{p^{j}_{i}}$, respectively. Consider, without loss of generality, the robots' positions and image projections expressed in unknown equally oriented frames common to all cameras. As $\boldsymbol{\rho_{i}^{j}}=\mathbf{p_{i}^{d,j}} - \mathbf{p_{i}^j}=r_{j}(\mathbf{x_{i}^{d,j}} - \mathbf{x_{i}})$, (\ref{eqrig}) can be expressed as:
\begin{equation}
\boldsymbol{\rho_{i}^{g}}=\overline{r}_i \sum_{j \in C_{i}} (\mathbf{x_{i}^{d,j}} - \mathbf{x_{i}}),
\label{emgi}
\end{equation}
\noindent where the factor --unknown to all aerial units and robots--
\begin{equation}
\overline{r}_i=\frac{1}{card(C_{i})} \sum_{k \in C_{i}} r_{k},
\end{equation}
\noindent is the average relative scale for the cameras that control robot $i$. Thus, the proposed scale adjustment in (\ref{eqrig}) makes the vectors $\boldsymbol{\rho_{i}^{j}}$ enter the computation of $\boldsymbol{\rho_{i}^{g}}$ with a consistent scale.

\vspace{1pt}
\noindent \textbf{Control law}: Robot $i$ computes (\ref{eqrig}) in its own frame, and:
\begin{eqnarray}
\label{erhomi}
\rho_{mi}=||\boldsymbol{\rho_{i}^{g}}||, \hspace{8pt} \alpha_{mi} = \pi - atan2 \left( (\boldsymbol{\rho_{i}^{g}})_{y}, (\boldsymbol{\rho_{i}^{g}})_{x} \right),
\end{eqnarray}
considering the robot's frame defined by its heading. The control goal for the robot is given by the ground position associated with the endpoint of its global motion vector. The variables $\alpha_{i}$ and $\rho_{i}$ (see Fig. \ref{figMHimageplane}) express this position. As $\alpha_{i}=\alpha_{mi}$ and $\rho_{i}$ is proportional to $\rho_{mi}$, we can control the robot using the image
quantities. The proposed control law for robot $i$ is:
\begin{equation}  \left\{ \begin{array}{l}
  v_{i} =  - k_{v} \, sign(\cos\alpha_{mi}) \, \rho_{mi}   \\
   \omega_{i} =  k_{\omega} \, (\alpha_{di} - \alpha_{mi})
\end{array} \right.
\label{eccontrol},
\end{equation}
\noindent where $k_{v}>0$ and $k_{\omega}>0$ are control gains, $\omega_{i}$ is considered in counterclockwise direction, and we define:
\begin{eqnarray} \nonumber
 \alpha_{di} = \left\{ \nonumber
 \begin{array}{cc} \nonumber
  0 & \textnormal{if} \quad |\alpha_{mi}| \leq \frac{\pi}{2} \nonumber
    \\ \nonumber
 \pi & \textnormal{if} \quad |\alpha_{mi}| > \frac{\pi}{2} \nonumber
\end{array} \nonumber
\right. . \nonumber
\nonumber
\end{eqnarray}
\noindent Angles are taken in $[-\pi,\pi)$. Observe that $0 \leq |\alpha_{mi}-\alpha_{di}| \leq \pi/2$  and that if $\cos\alpha_{mi}=0$, $v_{i}=0$ and robot $i$ can rotate in place but not translate. We define $\alpha_{mi}$ as $0$ if $\rho_{mi}=0$.
\begin{remark}
\label{rposoricams}
From (\ref{emgi}) and due to Property 1, a given robot's \textit{direction} of motion is independent from the cameras' locations and calibrations. Therefore, clearly, these factors do not affect the stability of the controller, studied in the following section. The height and calibration of the cameras influence the value of $\overline{r}_i$, having an effect equivalent to an unknown positive multiplicative gain acting on the linear velocity control (\ref{eccontrol}).
\end{remark}
\section{Stability analysis}
\label{sstabch5}
We study next the stability of the formation controller. We will consider common frames, only for analysis --recall that each UAV computes the control in its local image frame--. Consider the robots' ground positions $\mathbf{x_{i}}=[x_{i},y_{i}]^T$, $i=1,...,n$ expressed in an arbitrary global frame. We define the following cost function for the system under a topology $p \in \mathcal{P}$:
\begin{equation}
V=  \sum_{j=1,...,m}  V^{j},   \hspace{10pt} V^{j}= \frac{1}{2}\sum_{i \in S_{j}^{c}}  ||\mathbf{x_{i}^{d,j}}-\mathbf{x_{i}}||^2.
\label{ecostmulticam}
\end{equation}
Note that the state of the formation can be represented by the set of vectors $\mathbf{x_{i}^{d,j}}-\mathbf{x_{i}}$, $j=1,...,m$, $i \in S_{j}^{c}$, and $V$ is radially unbounded, as $||\mathbf{x_{i}^{d,j}}-\mathbf{x_{i}}||\rightarrow\infty$ for a pair $i,j$ implies $V\rightarrow\infty$. For generality, we use an alternative definition, common across all topologies, of the system's state, by defining the following stack state vector: $\mathbf{X}=[\mathbf{X_{1}}^T, \mathbf{X_{2}}^T, ..., \mathbf{X_{n}}^T]^T$ $\in \mathbb{R}^{2n}$, where $\mathbf{X_{i}}=\mathbf{x_{i}^{d}}- \mathbf{x_{i}}$, and $\mathbf{x_{i}^{d}}=[x_{i}^{d},y_{i}^{d}]^T$ is the ground position associated with the desired image point $\mathbf{p_{i}^{d}}$ obtained from any \textit{global} similarity (i.e., one computed from all the $n$ robots). Next, we establish two preliminary results:
\begin{lemma}
\label{lem_equil}
For any $p \in \mathcal{P}$, the robots form the desired shape if and only if $\mathbf{X}=\mathbf{0}$, which occurs if and only if $V=0$.
\end{lemma}
\begin{IEEEproof}{(sketch) If $V^j=0$ $\forall j$, we have all desired sub-shapes which, due to TC1, clearly fit together. When in the desired shape, desired and current points coincide, so $V=0$.}\end{IEEEproof}
\begin{lemma}
\label{lem_dvj}
For any topology $p \in \mathcal{P}$, it holds $\forall j=1,..,m$, $\forall i \in S_{j}^{c}$ that $\partial V^j  / \partial \mathbf{x_{i}}= \mathbf{x_{i}} - \mathbf{x_{i}^{d,j}}$.
\end{lemma}
\begin{IEEEproof}
Consider a camera $j$ and, without loss of generality, that the frames for $j's$ current image points, and the associated ground positions, are equally oriented and centered on their centroids, $\mathbf{c_{p}^{j}}$ and $\mathbf{c_{x}^{j}}$. The values $\mathbf{p'^{j}_{ic}}$ are projections of equivalent template ground positions $\mathbf{x'^{j}_{ic}}$ so, for $i \in S_{j}^c$:
\begin{equation}
\label{epointspositions}
\mathbf{p'^{j}_{ic}}=r_{j} \mathbf{x'^{j}_{ic}},\hspace{15pt}  \mathbf{p_{i}^{j}}=r_{j}
\mathbf{x_{i}}, \hspace{15pt} \mathbf{p^{d,j}_{i}} = r_{j} \mathbf{x^{d,j}_{i}}.
\end{equation}
\noindent As $\mathbf{p^{d,j}_{i}}=\mathbf{H_{s}^{j}}$ $\mathbf{p'^{j}_{ic}}$, we have
$\mathbf{x^{d,j}_{i}}=\mathbf{H_{s}^{j}}$ $\mathbf{x'^{j}_{ic}}$, and we can write:
\begin{equation}
V^j=\frac{1}{2}\sum_{i \in S_{j}^{c}} || \mathbf{x_{i}^{d,j}}-\mathbf{x_{i}}||^2=
\frac{1}{2}\sum_{i \in S_{j}^{c}}  || \mathbf{H_{s}^{j}} \mathbf{x'^{j}_{ic}} -\mathbf{x_{i}}||^2.
\label{elyap1cam}
\end{equation}
\noindent Clearly, if $\mathbf{H_{s}^{j}}$ fits, with least-squares error, the template ($\mathbf{p'^{j}_{ic}}$) and current ($\mathbf{p_{i}^{j}}$) image points (Section \ref{ssimila}), it also does so for the positions ($\mathbf{x'^{j}_{ic}}$, $\mathbf{x_{i}}$). As $V^{j}$ expresses precisely this sum of squared errors, $\mathbf{H_{s}^{j}}$ is the similarity that minimizes $V^{j}$. Considering
this transformation is unique and a differentiable function of the input points \cite{Gower04}, $\partial V^{j}/\partial {\mathbf{H_{s}^{j}}}$ is null. It is then direct that $\partial V^j  / \partial \mathbf{x_{i}}= \mathbf{x_{i}} - \mathbf{x_{i}^{d,j}}$ $\forall i \in S^{j}_{c}$, as claimed. 
\end{IEEEproof}
We now present the following main stability result:
\begin{theorem}
\label{theg}
For any fixed topology $p \in \mathcal{P}$, by using the control law (\ref{eccontrol}) the positions of the team of ground robots converge asymptotically to the desired formation shape.
\end{theorem}
\begin{IEEEproof}
We consider, without losing generality, that all ground positions are expressed in a common frame, with which all image frames are aligned. We take $V$ as a candidate Lyapunov function for the system. Its dynamics are:
\begin{equation}
\dot{V} =  \sum_{j=1}^{m} \left[\sum_{i \in S_{j}^c} \left(\frac{\partial V^{j}}{\partial \mathbf{x_{i}}}\right)^T {\mathbf{\dot{x}_{i}}} \right].
\label{dotlyapf}
\end{equation}
From (\ref{eccontrol}), (\ref{emgi}), and the unicycle kinematic model, we have:
\begin{equation}
{\mathbf{\dot{x}_{i}}}=k_{v} \mathbf{Q_i}\boldsymbol{\rho_{i}^{g}} =k_{v} \overline{r}_i \mathbf{Q_{i}}  \sum_{j \in C_{i}} (\mathbf{x_{i}^{d,j}} - \mathbf{x_{i}}),
\label{xdotmulti}
\end{equation}
\noindent where the misalignment between the robot's displacement
direction and $\boldsymbol{\rho_{i}^{g}}$ is captured by $\mathbf{Q_i} \in SO(2)$, a rotation
by the angle $\alpha_{mi}-\alpha_{di}$. Inserting (\ref{xdotmulti}) and Lemma \ref{lem_dvj} in (\ref{dotlyapf}) gives:
\begin{align}
&\dot{V}  = \sum_{j=1}^{m} \Bigg\{\sum_{i \in S_{j}^c} \Bigg[(\mathbf{x_{i}} - \mathbf{x_{i}^{d,j}})^T \Big(k_{v} \overline{r}_{i} \mathbf{Q_{i}} \sum_{j \in C_{i}}(\mathbf{x_{i}^{d,j}}- \mathbf{x_{i}})\Big) \Bigg] \Bigg\}
\nonumber \\
&= -k_{v} \sum_{i=1}^{n} \overline{r}_{i} \cos(\alpha_{mi}-\alpha_{di}) || \sum_{j \in C_{i}} (\mathbf{x_{i}^{d,j}}- \mathbf{x_{i}})||^2 \leq 0,
\label{dotvmultiddd}
\end{align}
\noindent where the inequality holds as $0 \leq |\alpha_{mi}-\alpha_{di}| \leq \pi/2$. From the invariant set theorem, the system is locally stable with respect to $V=0$, i.e., the desired team shape (Lemma \ref{lem_equil}). \\
We can guarantee global asymptotic stability, if the only equilibrium (i.e., $\dot{V}=0$) of the system occurs at $\mathbf{X}=\mathbf{0}$ (i.e., $V=0$). Due to the unicycle kinematics, $\dot{V}$ may be zero if no robot is translating, and at least one of them satisfies $\cos(\alpha_{mi}-\alpha_{di})=0$ and $||\mathbf{\rho_i^g}|| > 0$. However, these robots will rotate in place at that moment, immediately making $\dot{V}<0$. Hence, the only relevant scenario to examine is $||\mathbf{\rho_i^g}|| = 0, \hspace{3pt} i=1,...,n$. Using the topological conditions TC1-TC5 and the constraints they impose on the robots' motion vectors, and via a similar analysis to the one presented in the proof of \cite[Corollary~1]{ArandaTRO15}, one can see through simple geometric conditions that if a sum vector $\mathbf{\rho_i^g}$ is null, the individual vectors $\mathbf{\rho_i^j}$ must be null, too. Thus, the only possible stable equilibrium occurs at $V=0$, i.e., the team of robots converges asymptotically to the prescribed shape. 
\end{IEEEproof}
\begin{corollary}
\label{cor2}
It is direct to see that the robots remain static once the desired shape has been achieved.
\end{corollary}
\begin{corollary}
\label{cor1}
The distance between every two robots remains upper-bounded with control law (\ref{eccontrol}), for any topology $p \in \mathcal{P}$.
\end{corollary}
\begin{IEEEproof}
\noindent For finite initial robot positions, $V$ (\ref{ecostmulticam}) is clearly upper-bounded and, as $\dot{V}\leq0$ (Theorem \ref{theg}) it remains so for all time. As $V$ is the sum of squared norms of the vectors $\mathbf{x_{i}}-\mathbf{x_{i}^{d,j}}$, all these norms are also always upper-bounded, and so are the norms of $\mathbf{\rho_{i}^{j}}$ (\ref{erhoij}) and, therefore, the norms of the motion vectors $\mathbf{\rho_{i}^{g}}$ (\ref{eqrig}). Therefore, the magnitudes of the linear velocities for all robots (\ref{eccontrol}) are also always upper-bounded, and the distance between any two robots may only become unbounded in infinite time. We can now consider, for all practical purposes, an arbitrarily small positive threshold $b_{th}$ that stops the robots' motions (i.e., $\forall i$, $v_{i}=\omega_{i}=0$ if $\rho_{mi}<b_{th}$). Then, since all robots will clearly --due to the vanishing behavior of $V$-- stop displacing in finite time, the inter-robot distances will be upper-bounded. 
\end{IEEEproof}
\begin{corollary}
\label{corlocal}
For $p \in \mathcal{Q}$, clearly, Lemma \ref{lem_equil} holds and, thanks to Theorem \ref{theg}, the formation controller is locally stable.
\end{corollary}
\begin{corollary}
\label{coronecam}
For the particular case $m=1$, $S_{1}^{c}=S$ (i.e., a single UAV controls all the ground robots), it is direct from Theorem \ref{theg} that the formation controller is globally convergent.
\end{corollary}
\begin{remark}
The control may have singularities for certain robot arrangements that are non-attracting and have zero measure \cite{Gower04,ArandaTRO15}. Thus, for all practical purposes, $\mathbf{H_{s}^{j}}$ is differentiable (Lemma \ref{lem_dvj}) and the system stable. Alternatively, we could consider the degenerate cases and use an \textit{almost-global} stability result. Also, even if control law (\ref{eccontrol}) is discontinuous, the vanishing behavior of $V$ suffices to prove stability. As the angular velocity in (\ref{eccontrol}) always drives the system \textit{away from} these discontinuities, chattering-like behaviors are not feasible.
\end{remark}
\subsection{Stability with changes in topology}
Due to the motion of UAVs and ground robots during execution, the latter may come in and out of the fields of view of the cameras, so the sets $S_{j}$ and $S^{c}_{j}$ will switch. This affects $\mathbf{H_{s}^{j}}$ (\ref{eHeuclideanl}) and in turn, via $\mathbf{p_i^{d,j}}$ (Algorithm \ref{algorithmcam}) and (\ref{erhoij}), (\ref{eqrig}), (\ref{erhomi}), the control law (\ref{eccontrol}). Thus, due to the topology changes, ours is a \textit{switched system} \cite{Liberzon03}, which we analyze as follows.
\begin{proposition}
Consider that the controller switches within the possible topologies $p \in \mathcal{P}$. Then, there exists a finite positive value $\tau_d$ such that the team of robots under control law (\ref{eccontrol}) converges asymptotically to the formation shape if the average dwell time of every topology is at least $\tau_d$.
\end{proposition}
\begin{IEEEproof}
$\mathcal{P}$ is a finite set, and the system's state (determined by the agents' positions) does
not jump at switching times. Also, the dwell time of every topology
is lower-bounded by a positive value (Section \ref{ssysdes}). From Theorem \ref{theg}, the formation is
asymptotically stable for every individual $p$. All topologies drive the system to the same common
equilibrium ($\mathbf{X}=\mathbf{0}$), but each has a different Lyapunov function --i.e., $V_{p}$ for topology $p$--. Thus, from \cite{Liberzon03,Hespanha99}, there is a finite average dwell time that guarantees asymptotic stability if:
\begin{equation}
V_{p}(\mathbf{X}) \leq \mu V_{q}(\mathbf{X}) \hspace{10pt} \forall \mathbf{X} \in \mathbb{R}^{2n}, \hspace{2pt}\forall p, q \in \mathcal{P},
\label{lyapdwell}
\end{equation}
\noindent for a given constant $\mu$. The Lyapunov function for a given topology consists of the sum of
squared metric distances. Let us call these distances $d_{i}^{j}=||\mathbf{x_{i}^{d,j}}-\mathbf{x_{i}}||$. If
$\mathbf{X} = \mathbf{0}$, all $d_{i}^{j}=0$ for all topologies. Otherwise, there is at least one $d_{i}^{j}>0$ for
every topology. Denote as $d_{i}^{j*}$ those $d_{i}^{j}$ that are strictly positive. From Corollary \ref{cor1}, for
all topologies, inter-robot distances remain upper-bounded; therefore, the desired
positions $\mathbf{x_{i}^{d,j}}$ are such that all $d_{i}^{j*}$ are always upper-bounded, too. Therefore, the ratio of
any two $d_{i}^{j*}$ is bounded --i.e., we can define a finite value $B=max_{i,j,p,\mathbf{X}}(d_{i}^{j*}) / min_{i,j,p,\mathbf{X}}(d_{i}^{j*})$--. Now, as each Lyapunov function is the sum of a finite number
of $d_{i}^{j}$ distances squared, it follows that a finite $\mu$ in (\ref{lyapdwell}) must exist. Hence, the statement of the Proposition holds true. 
\end{IEEEproof}
This result means that if every topology is active, on average, for a sufficiently long time, the desired ground team shape will be attained. Other interesting properties are that the switching is controllable --the aerial units can, through their coordinated motions and decisions, determine the switches-- and typically will stop in finite time --clearly, when the ground team is close to the prescribed shape, no topology switches are needed--.
\section{Motion and coordination of the aerial units}
\label{scoordaerial}
We give next guidelines to implement the aspects of UAV control and coordination, whose detailed study is not the focus of this paper. A key observation is that high-speed and precise aerial unit motions are not required: the UAVs do not need to react fast or reach specific positions, the control is inherently robust to imperfect UAV motions (Remark \ref{rposoricams}), and we can define safety margins to aid their maneuverability. Thus, it is reasonable to model the UAV translational motion at a kinematic --and not dynamic-- control level (see, e.g., \cite{Bourquardez09}). For simplicity, in our tests we use single-integrator kinematics.

In terms of their coordination, we propose to make the UAVs follow the algorithm in \cite{ArandaTRO15}, which exploits communications (Section \ref{ssysdes}) of image data and ensures TC1-TC2 by preserving the links of the initial graph $\mathcal{G}_{c}(t=0)$. To initially deploy the UAVs without full knowledge of the ground robot locations, distributed coverage/search algorithms with connectivity maintenance features can be used. A simple approach can be, e.g., to deploy the UAVs one-by-one sequentially while enforcing TC1-TC2, and create an initial path graph $\mathcal{G}_{c}$. Note that each UAV controls its displacement to preserve in the field of view the \textit{two} closest robots seen in common with each one of its neighbors in $\mathcal{G}_{c}$; thus, additional common robots can leave the UAV's control scope as the system evolves (i.e., transfers of robots between UAVs are possible). TC1-TC5 can be met by suitably defining $S_{j}^c$ (which are subsets of $S_{j}$, $\forall j$) via distributed protocols, implemented for a duo of neighboring UAVs using, e.g., image distance-based criteria. The duo can thus decide which two of their common viewed ground robots they will share the control over, and which of the two UAVs will assume (if needed) control of the other common viewed robots. The activation time of each topology can also be ensured to be lower bounded. The exchanged data ($S_{j}$ and image points) between neighboring UAVs could be used in more efficient and flexible coordination schemes to, e.g., balance the load (i.e., cardinality of sets $S_{j}^c$), and recover the affected ground robots if a UAV fails.
\begin{figure}
\centerline{
\begin{tabular}{c}
\includegraphics[trim=1cm 18pt 1cm -2pt, width=0.62\linewidth]{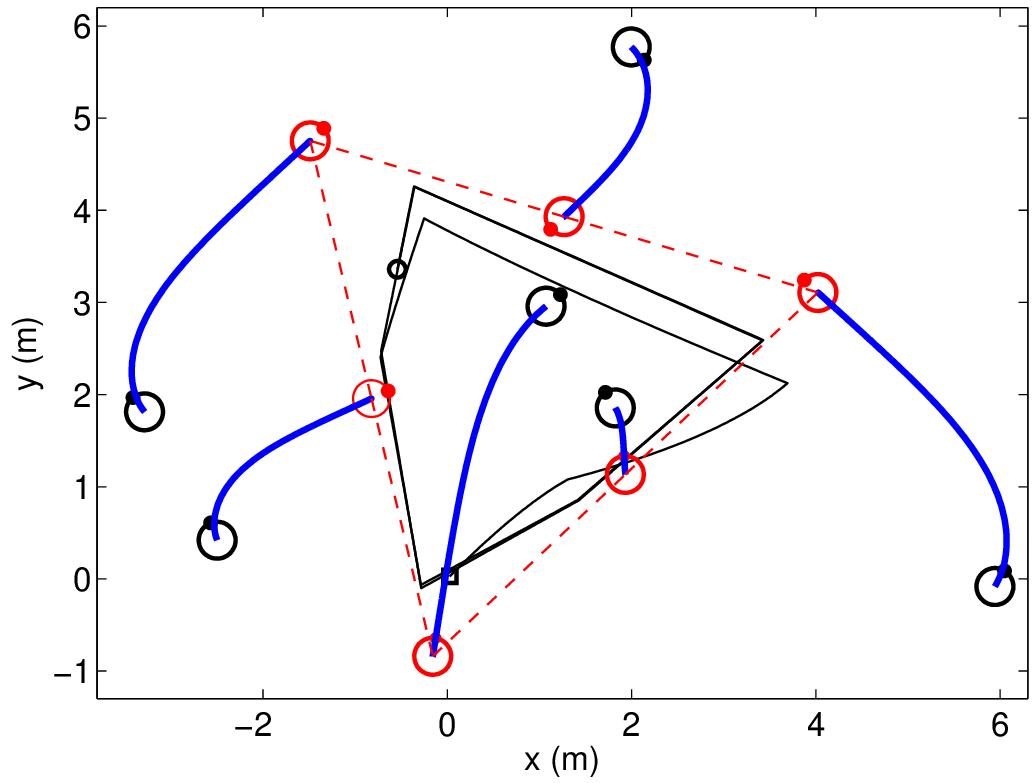} \\ 
\includegraphics[trim=0 20pt 0 0cm, width=0.46\linewidth]{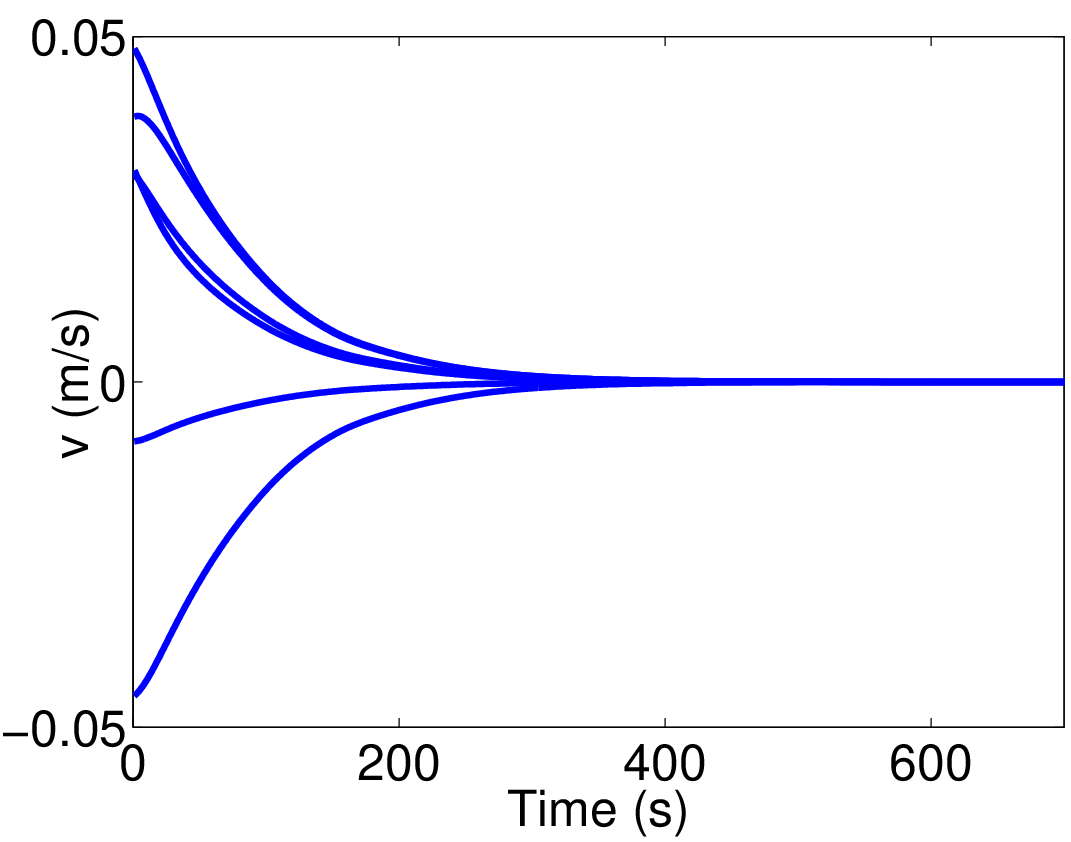}
\includegraphics[trim=0 20pt 0 0cm, width=0.435\linewidth]{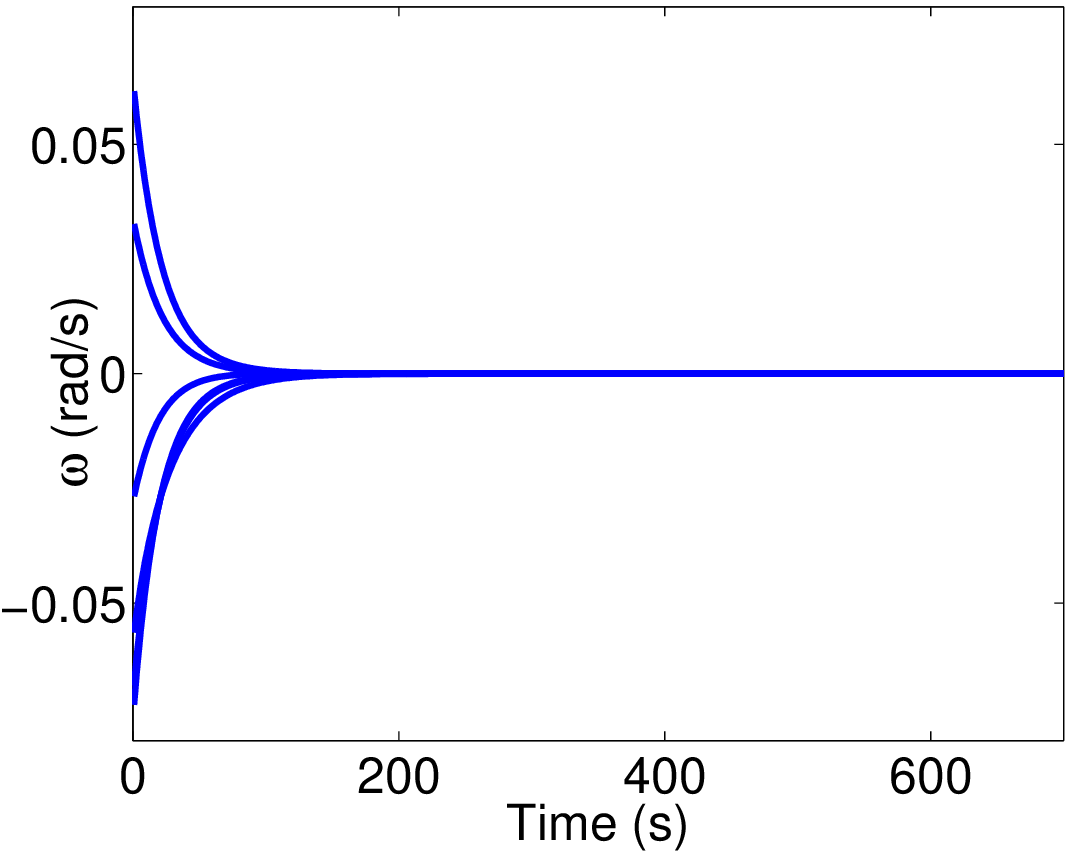} \\
\includegraphics[trim=0 12pt 0 0cm, width=0.472\linewidth]{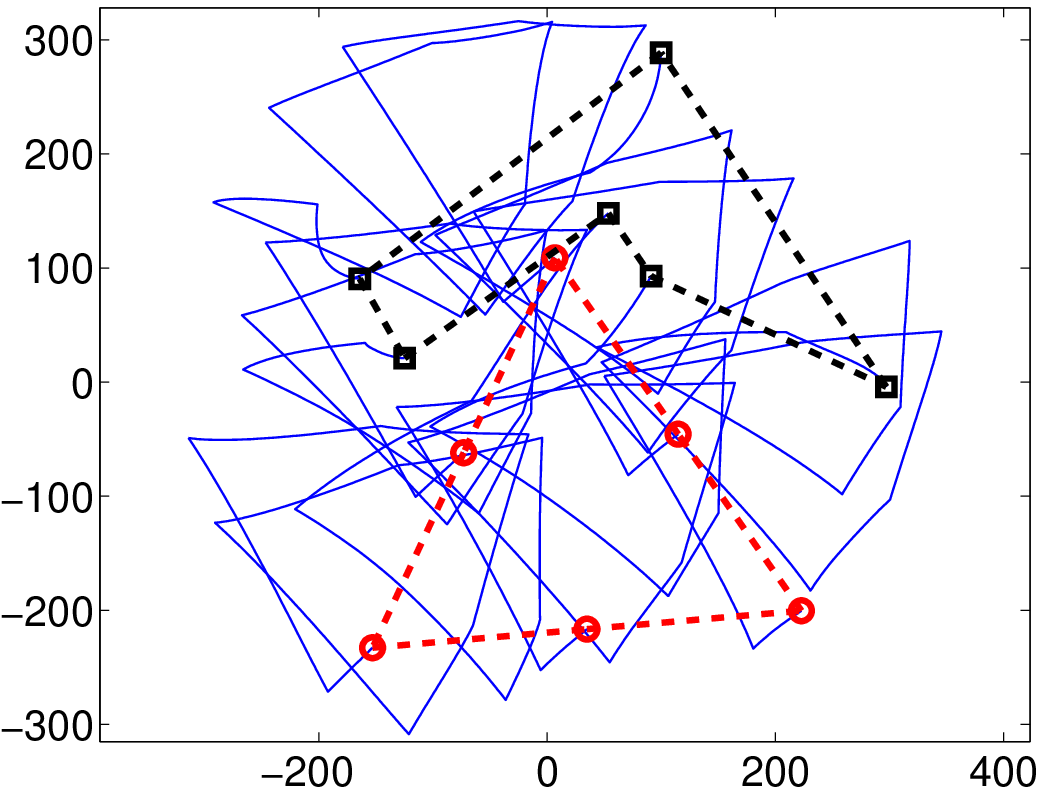}
\includegraphics[trim=0 12pt 0 0cm, width=0.425\linewidth]{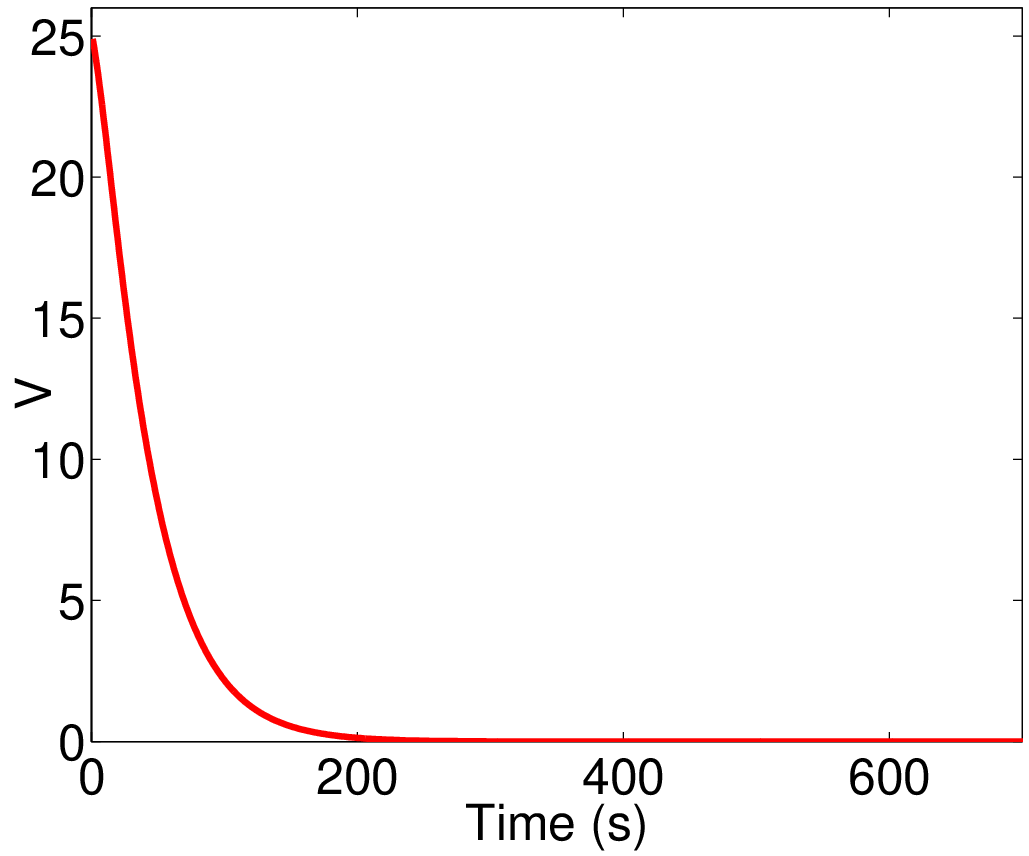} \\
\includegraphics[trim=0 26pt 0 0cm, width=0.445\linewidth]{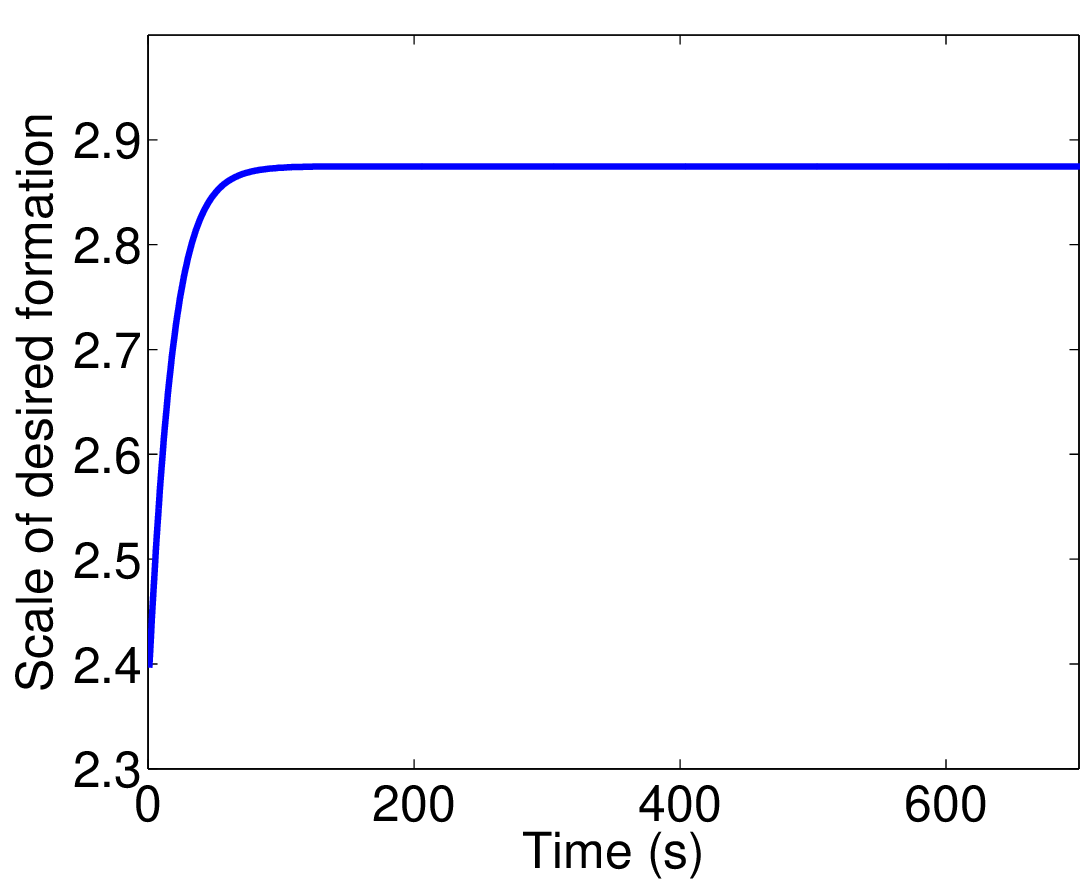}
\includegraphics[trim=0 26pt 0 0cm, width=0.472\linewidth]{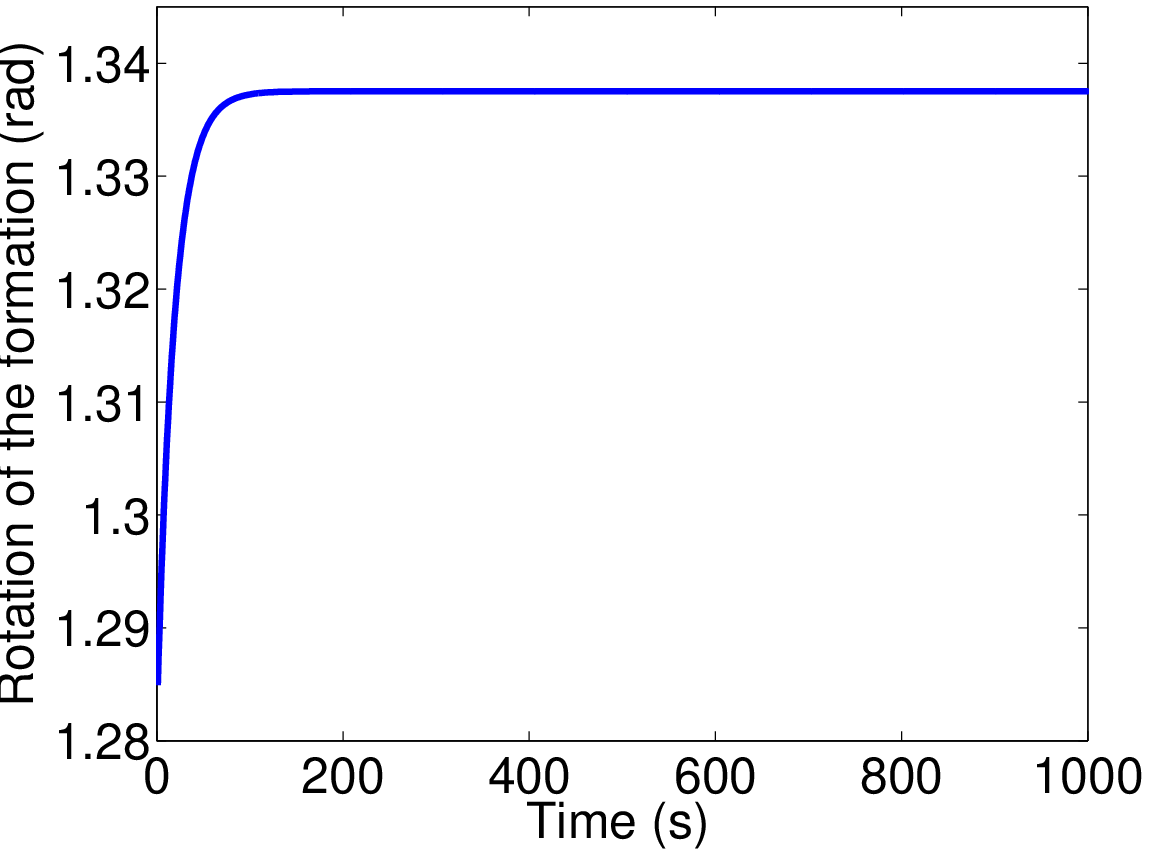} \\
\includegraphics[trim=0 28pt 0 0cm, width=0.455\linewidth]{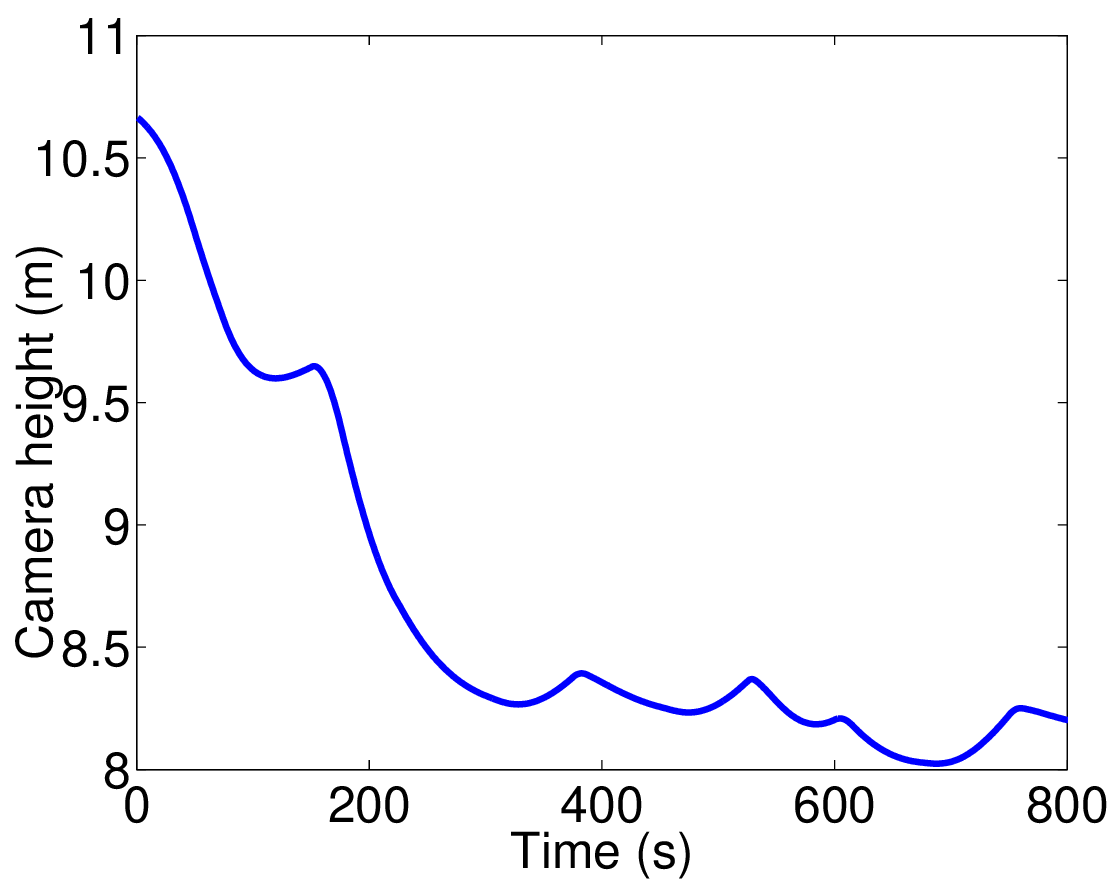}
\includegraphics[trim=0 28pt 0 0cm, width=0.455\linewidth]{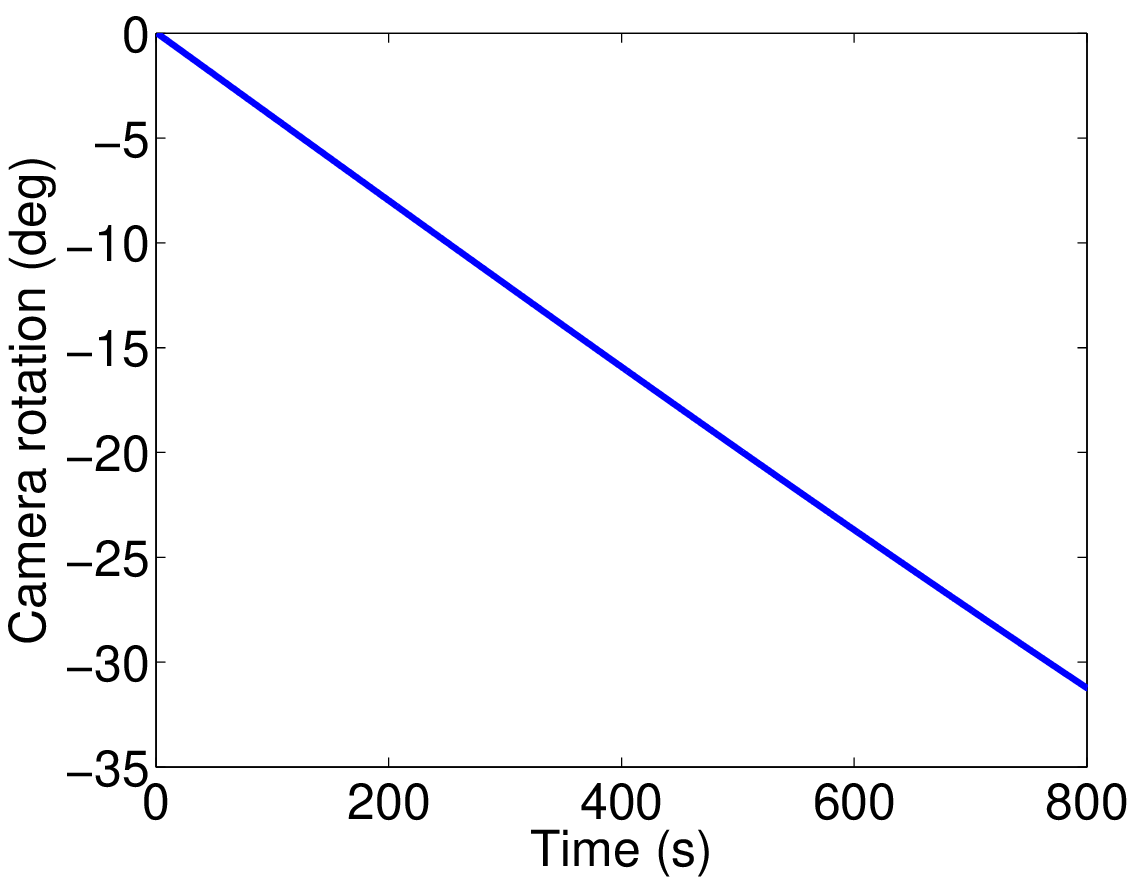}
\end{tabular}
} \caption{One-camera simulation results. Top: robot paths (final positions joined by dashed lines) and projection of camera path on ground plane (initial point marked as square, final as circle). Row 2: linear (left), angular (right) robot velocities. Row 3: image traces of robots --initial and final point sets joined by dashed lines-- (left) and cost function (right). Row 4: scale (left) and rotation (right) of desired formation. Row 5: camera height (left) and rotation (right).}\label{fig_sim_1cam}
\end{figure}
\section{Simulation results}
\label{ssimaer}
In this section, we illustrate the performance of the control
method with simulations. We first describe an example where one downward-facing camera was used to drive a group of six unicycle robots to a triangular desired shape. Figure \ref{fig_sim_1cam} displays the results, showing how the formation was achieved. The aerial unit displaced horizontally following the perimeter of the ground formation. By doing so --instead of, e.g., remaining over the team's centroid--, it can gain a richer perception of the ground team's surroundings so as to, e.g., detect obstacles or threats. Note that this persistent UAV motion does not affect the formation's convergence. The camera always maintained a good visibility of all the robots, as seen in the image traces. The UAV controlled its vertical motion to guarantee this visual coverage, and it also rotated during execution. We stress that the UAV used only image information, expressed in pixels.

\begin{figure*}
\centerline{
\begin{tabular}{c}
\begin{tabular}{cc}
\begin{tabular}{c}
\includegraphics[trim= 1cm 0cm 1cm 0, width=0.33\linewidth]{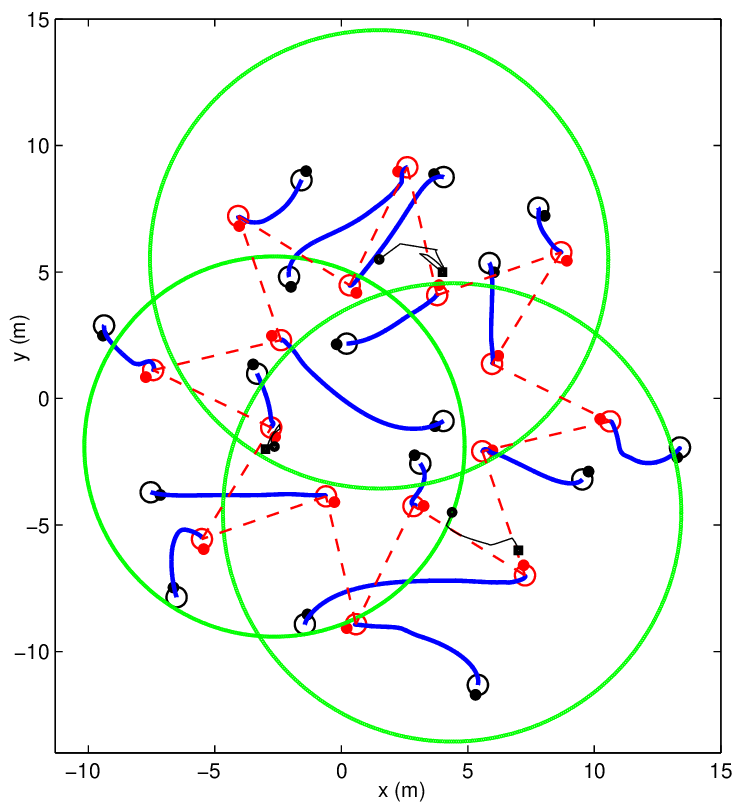}
\end{tabular}
&
\begin{tabular} {c}
\includegraphics[width=0.26\linewidth]{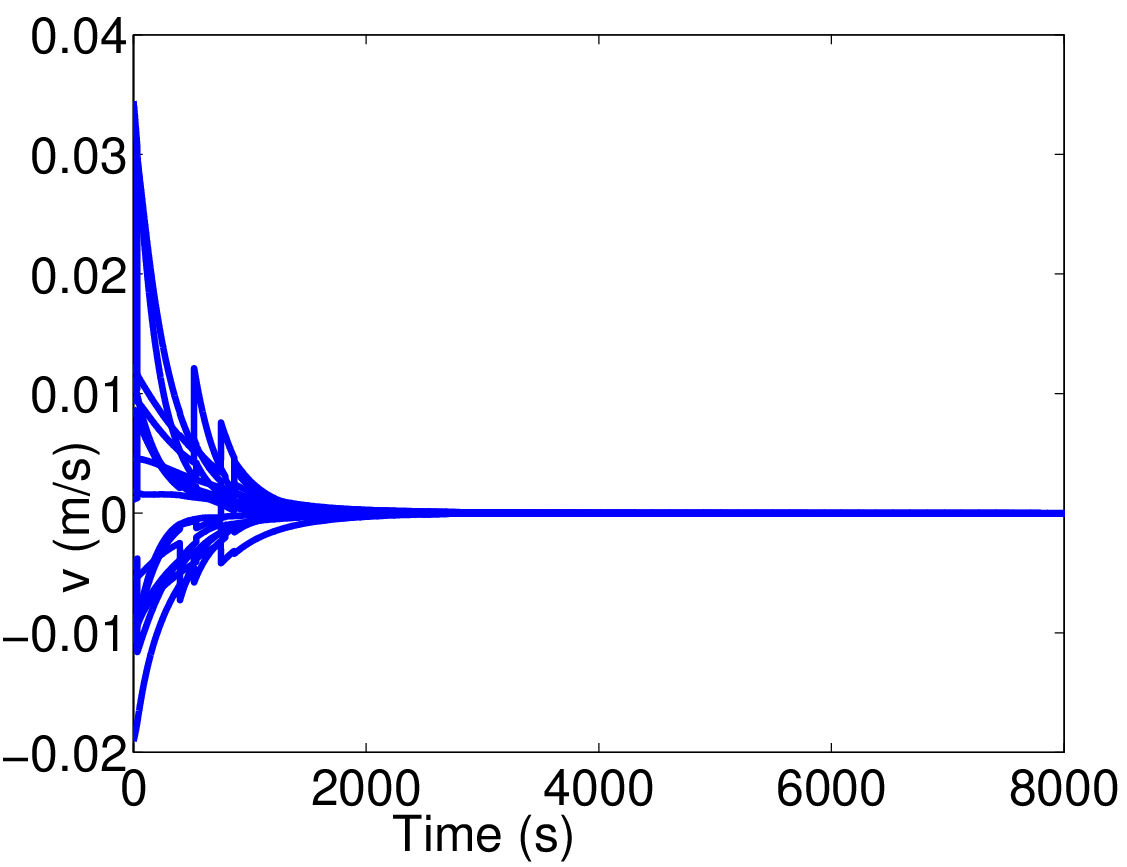}
\includegraphics[width=0.257\linewidth]{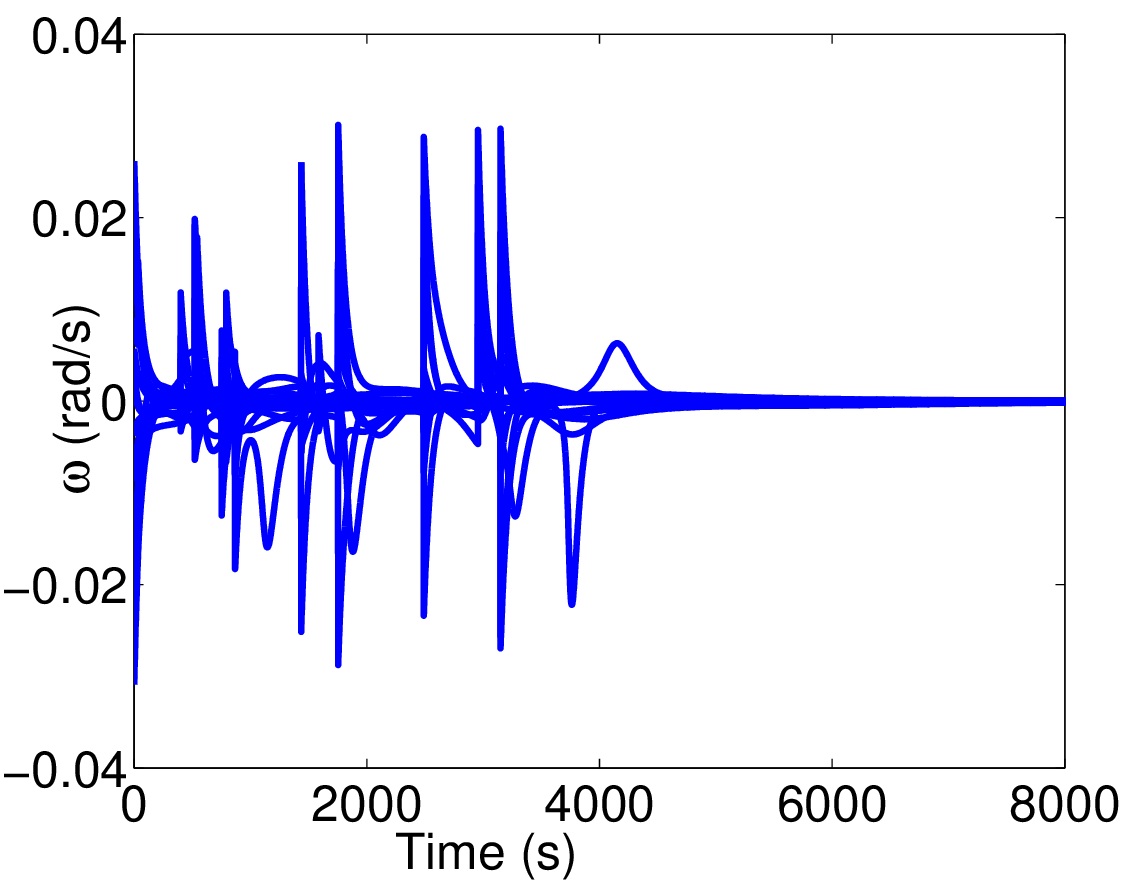}\\
\includegraphics[width=0.2613\linewidth]{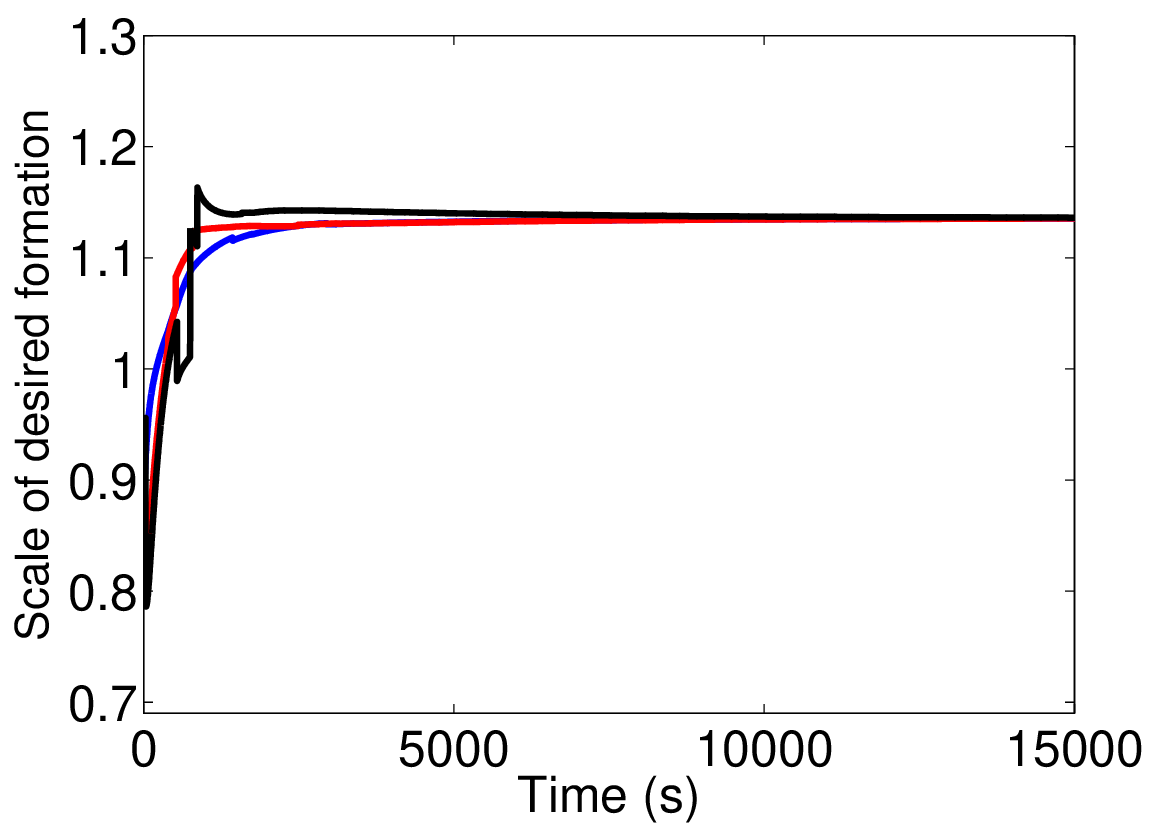}
\includegraphics[width=0.2675\linewidth]{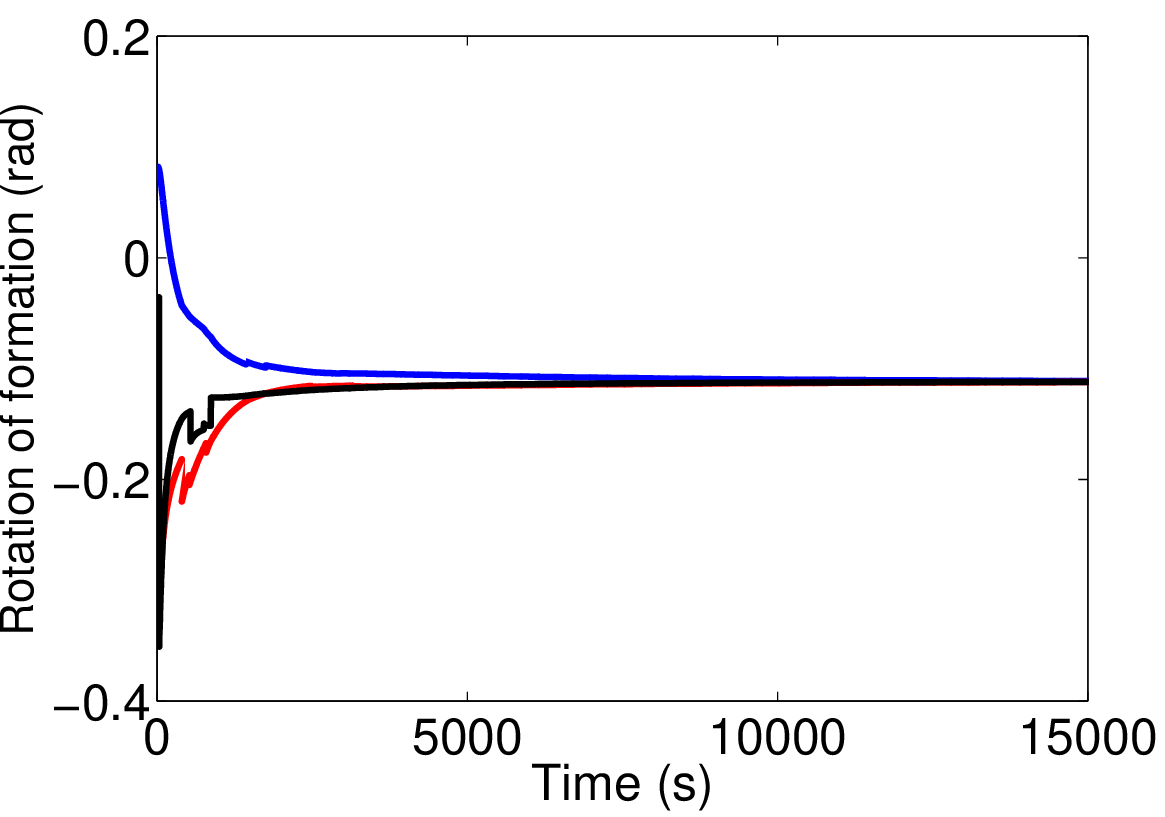}
\end{tabular}
\end{tabular}
\\
\includegraphics[width=0.2155\linewidth]{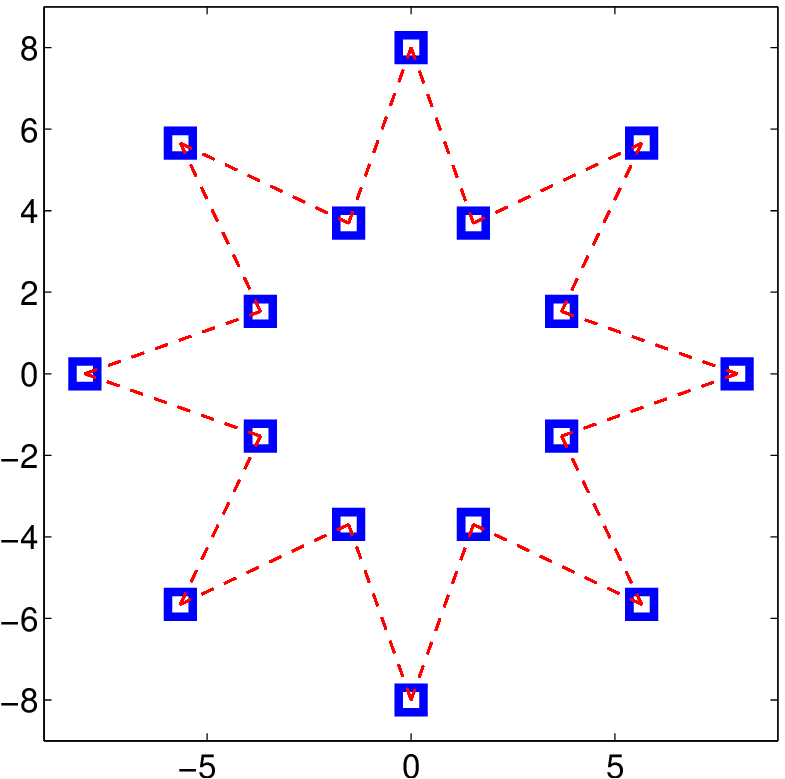}
\includegraphics[width=0.28\linewidth]{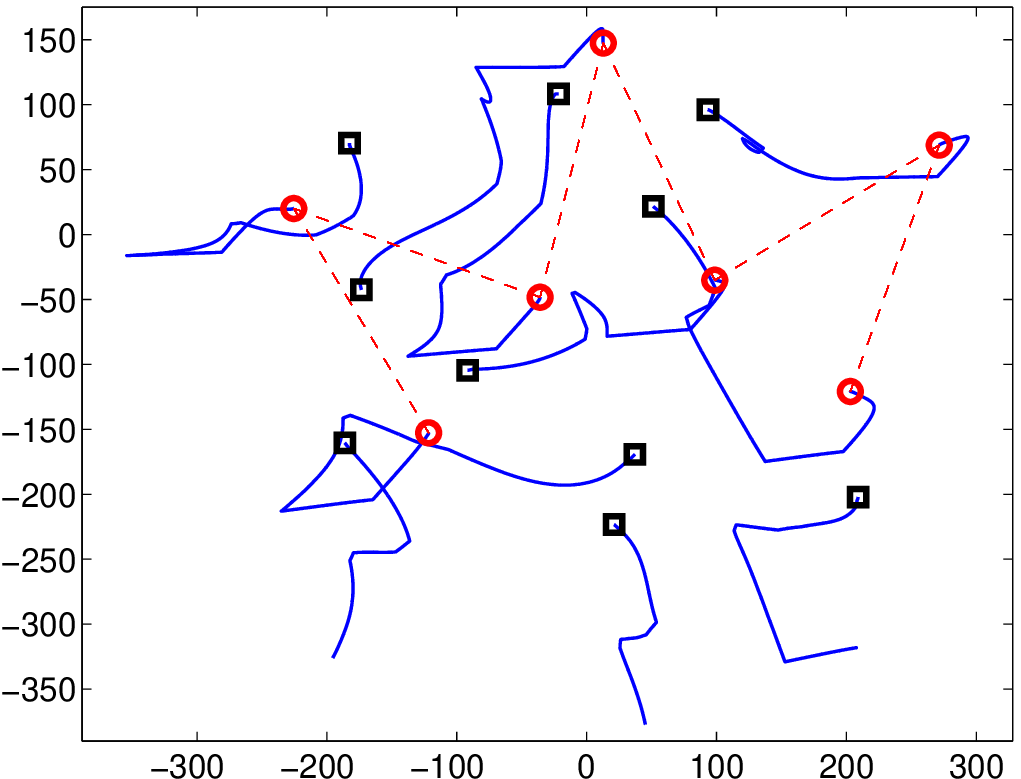}
\includegraphics[width=0.25\linewidth]{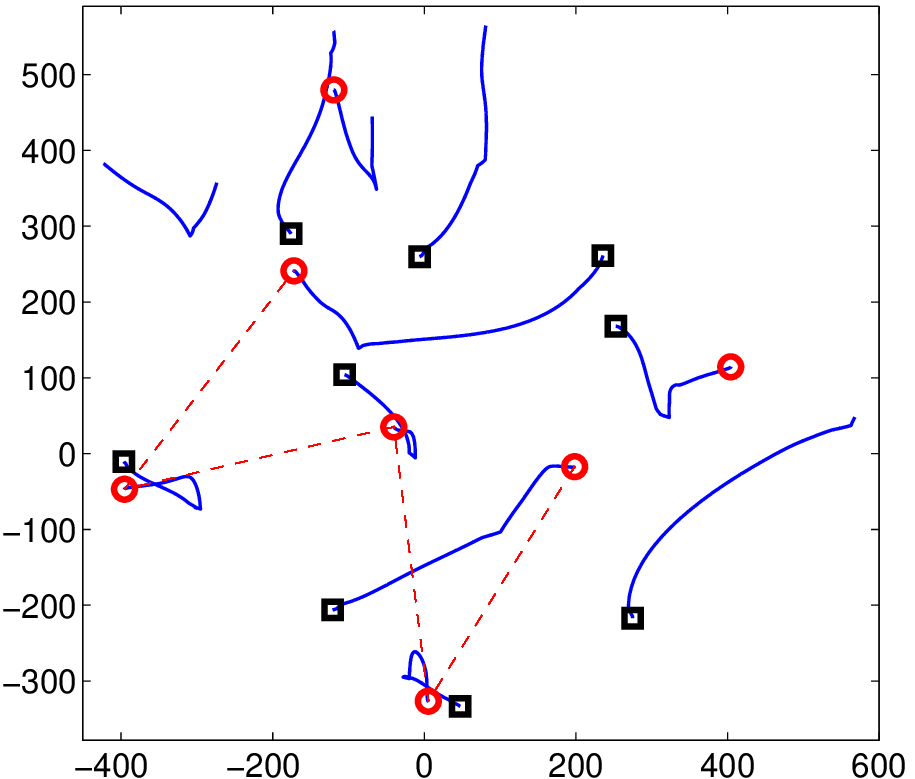}
\includegraphics[width=0.21\linewidth]{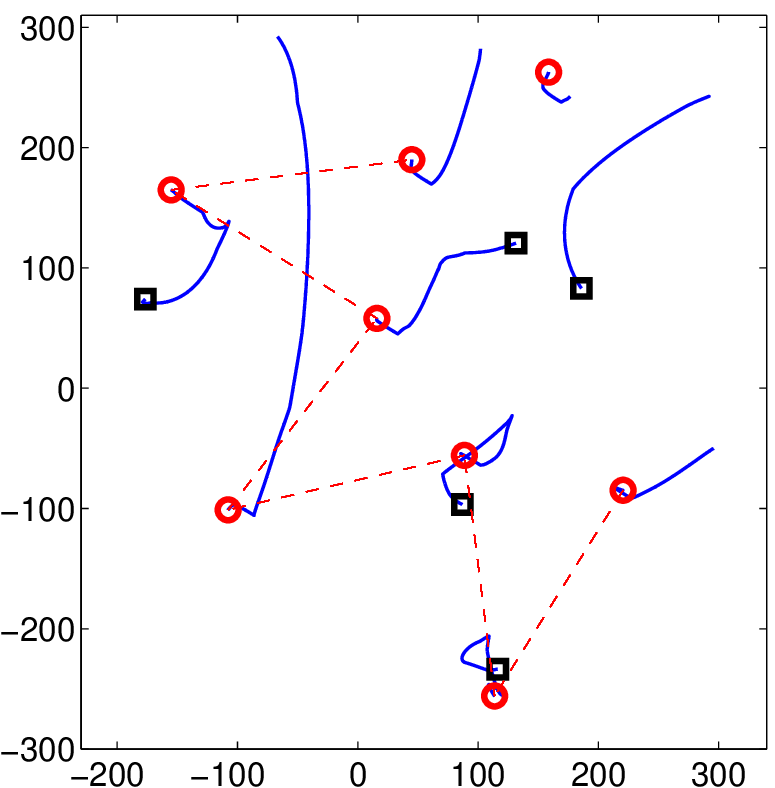}\\
\includegraphics[trim= 0 0 0 160pt, width=0.26\linewidth]{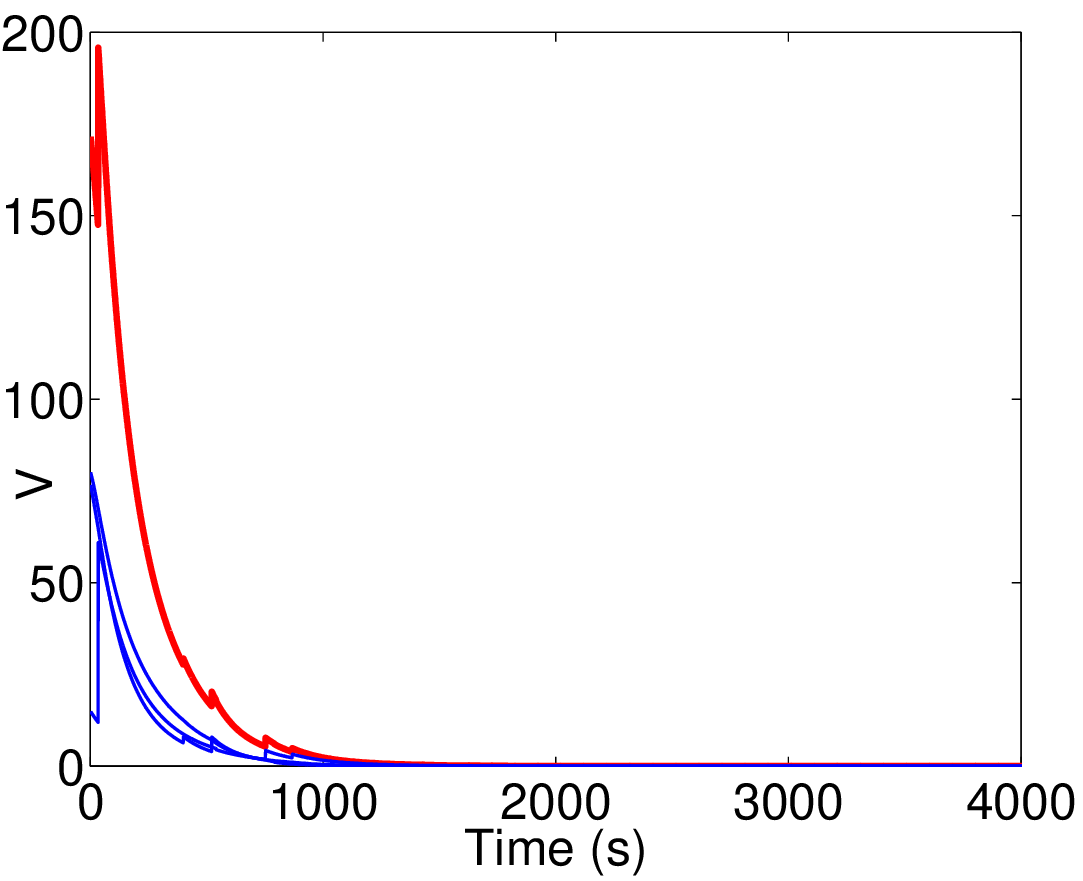}
\includegraphics[width=0.289\linewidth]{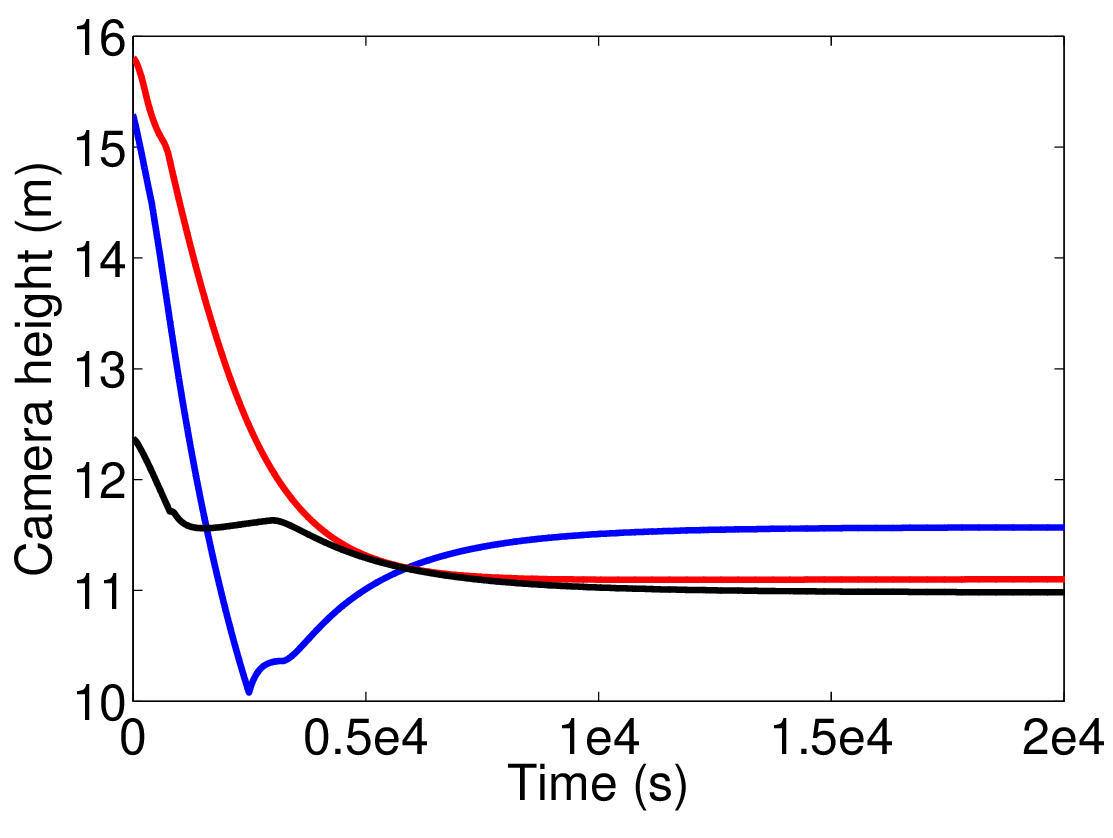}
\includegraphics[width=0.26945\linewidth]{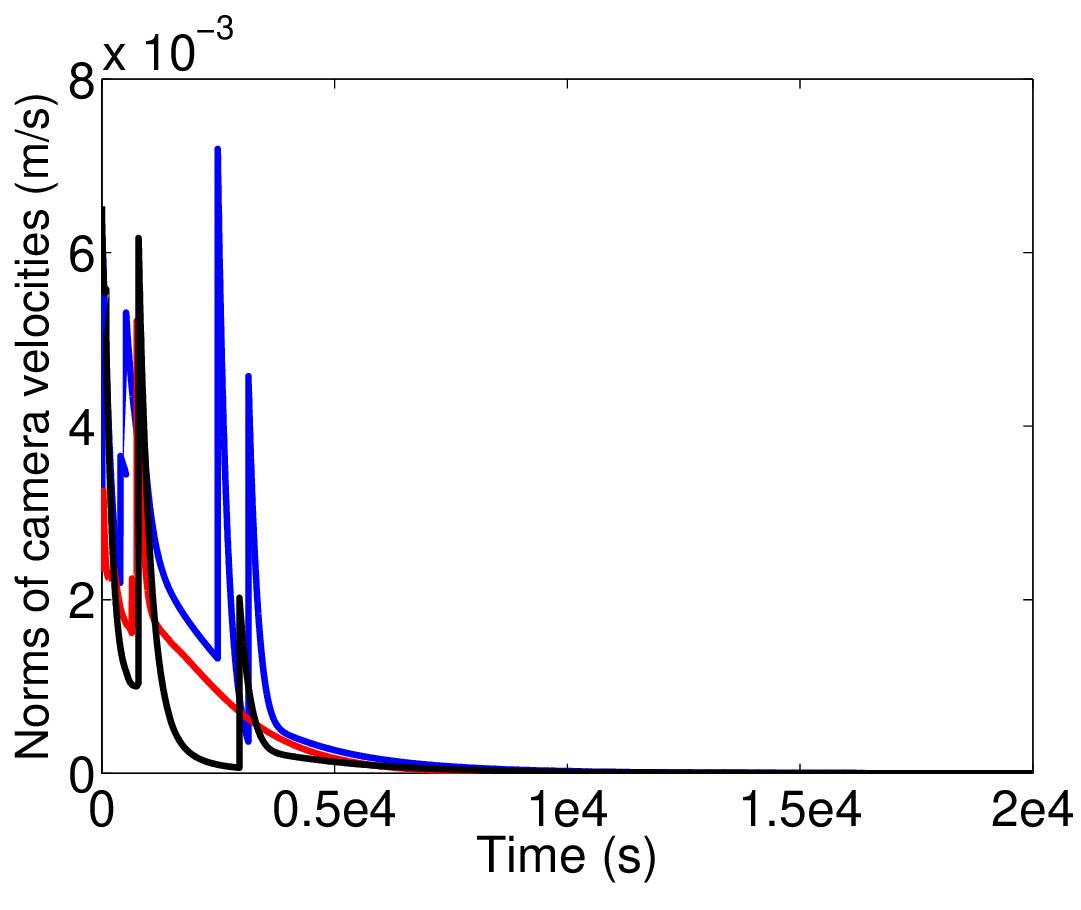}
\end{tabular}
} \caption{Simulation results. Top-left: Robot paths --final positions joined by dashed lines--, final camera
fields of view --circles--, and paths of the three cameras; Top-right panel: evolution of linear and
angular velocities of the ground robots (top), scales and rotations of the partial desired formations (bottom). Second row: template image (left), and image traces of the robots in $S_{j}^{c}$ for the three cameras (initial points marked as squares, final points as circles, joined by dashed lines). Bottom row: evolution of $V$ (highest-valued curve) and the three $V^{j}$ (left), camera heights (center) and camera velocity norms (right).}\label{fig_sim3cams}
\end{figure*}

In another example, three aerial units controlled a team of sixteen robots, to make them form a star-shaped configuration. Throughout the simulation, the UAVs moved as discussed in Section \ref{scoordaerial}, and generated ground control commands as described in Section \ref{smulti}. Each unit $j$ controlled those robots in $S_j$ closer than certain safety thresholds to the center of the image. The three cameras had different calibration. The UAVs only translated and did not rotate. Each had a different orientation. The results in Fig. \ref{fig_sim3cams} illustrate how the desired ground shape was achieved. There were topology switches, which caused the discontinuous changes observable in the plots. The cameras eventually stabilized to fixed positions. The plots show the scales and rotations of the \textit{partial} formations controlled by the UAVs, expressed in an arbitrary common fixed reference unknown to any UAV (note that these are \textit{not} the scales and rotations of the image similarities (\ref{eHeuclideanl}), which remain different for each camera and change continuously as they move). As expected, the three scales and rotations end up being equal as the team acquires the prescribed shape.
\section{Practical Discussion and Conclusion}
\label{sdiscussion}
We first discuss some application details and advantages and then limitations and potential improvements of our method.\\
\noindent $\bullet$ The proposed multi-UAV hybrid topology is scalable --it can include an arbitrarily large number of ground robots, and workspace-- and reliable --there is no central point of failure--.\\
\noindent $\bullet$ The ground robots are freed from sensing, costly processing and wireless transmission. Thus, one can use simple, low cost robots which will also have higher autonomy due to reduced power consumption. Their resources can hence be focused on other tasks (e.g., environmental monitoring or exploration).\\
\noindent $\bullet$ The UAVs do not have to achieve specific relative positioning or to synchronize their orientations. Thus, they are free to consider other concurrent goals --aside from ground formation achievement-- and, hence, fully exploit the known advantages of heterogeneous air/ground teams \cite{Lacroix2011,Chen2016}: they can monitor and preserve the system connectivity, reconfigure in case of failures, and their rich aerial imaging can enhance the navigation capabilities of the ground robots.\\
\noindent $\bullet$ As it does not need camera calibration and knowledge of scene scale, our method is robust to calibration errors and drifts --which are known to affect visual control stability--, allows to mount or change the cameras without preparative procedures, and, e.g., directly allows the use of zoom --which is clearly a very powerful feature for the scenario and task we consider--.\\
\noindent $\bullet$ Clearly, the UAVs will need to have localization information to enable them to navigate, which can be available in the infrastructure-free scenario we consider via, e.g., existing visual-inertial approaches. Our ground control is robust to errors in this information as it does not employ it.\\
\noindent $\bullet$ By using only \textit{local} (image) measurements, our method avoids the issues associated with using a shared UAV reference frame: need to maintain the agreed frame (requiring consensus or synchronization), inaccuracies in its definition --and their propagation among UAVs--, or inter-UAV communication issues (temporary losses, multi-hop delays...).\\
\indent To summarize, our method demands only simple resources from the ground robots and does not need a complex coordination strategy for the UAVs, provides a flexible architecture, and has useful decoupling properties and robustness to various typical sources of error. All this facilitates simpler implementation and integration of other tasks (aerial and ground perception/actuation) with the formation control itself.\\
\noindent \textit{Limitations and possible improvements:} Although the two-layer architecture provides distribution, the failure of a UAV affects not one but multiple robots --until other UAVs recover from it--. Also, it can be hard to visually detect and identify all robots in challenging conditions. Using interchangeable robots could be more robust and efficient, at the cost of more complex coordination. Performance will be perturbed if UAV disturbances make the camera not face downward --although image rectification as in \cite{Lopez-Nicolas12} can mitigate this-- or there are terrain irregularities. Finally, collision avoidance --e.g., via reactive methods-- for robots and UAVs should be used.
\bibliographystyle{./IEEEtran}

\end{document}